\documentclass[runningheads]{llncs}
\usepackage{graphicx}

\usepackage{tikz}
\usepackage{comment}
\usepackage{amsmath,amssymb} 
\usepackage{color}
\usepackage{caption}
\usepackage{subcaption}
\usepackage{booktabs}
\usepackage{comment}
\usepackage{epsfig}
\usepackage{mathtools}
\usepackage{breqn}
\usepackage{multirow}
\usepackage{adjustbox}
\usepackage{rotating}
\usepackage{diagbox}
\usepackage[colorlinks=true]{hyperref}

\usepackage[accsupp]{axessibility}  
\usepackage{orcidlink}

\newcommand{\etal}{\textit{et al.}}

\begin{document}
\pagestyle{headings}
\mainmatter
\def\ECCVSubNumber{6108}  

\title{Vote from the Center: 6 DoF Pose Estimation in RGB-D Images by Radial Keypoint Voting} 

\titlerunning{Vote from the Center: 6 DoF Pose Estimation in RGB-D Images}
%

\author{Yangzheng Wu\orcidlink{0000-0001-8893-0672}  \and Mohsen Zand\orcidlink{0000-0001-8177-6000
} \and
Ali Etemad\orcidlink{0000-0001-7128-0220} \and
Michael Greenspan\orcidlink{0000-0001-6054-8770}}
\authorrunning{Y. Wu et al.}
%
\institute{Dept. of Electrical and Computer Engineering, 
Ingenuity Labs Research Institute \\ Queen's University, Kingston, Ontario, Canada }
\maketitle

\begin{abstract}
We propose a novel keypoint 
voting scheme
based on intersecting
spheres,
that is 
more accurate
than existing schemes
and allows for 
fewer, more disperse
keypoints.
The scheme is based upon the
distance between points, which as a 1D quantity
can be regressed more accurately than the 2D and 3D vector and offset quantities regressed in previous work, yielding more accurate
keypoint localization.
The scheme forms the basis of 
the proposed RCVPose method for 6 DoF pose estimation of 3D objects in RGB-D data,
which is particularly effective at handling occlusions. A CNN is trained to estimate the distance between the 3D point corresponding to the depth mode of each RGB pixel, and a set of 3
disperse keypoints defined in the object frame. At inference, a sphere 
centered at each 3D point
is generated, 
of radius equal to this estimated distance. The surfaces of these spheres vote to increment a 3D accumulator space, the peaks of which indicate keypoint locations. 
The proposed radial voting scheme is more accurate than previous vector or offset schemes, and is robust to disperse keypoints. 
Experiments 
demonstrate RCVPose to be highly accurate and competitive,
achieving state-of-the-art results 
on the LINEMOD ($99.7\%$) and 
YCB-Video
($97.2\%$) datasets,
notably scoring
$+4.9\%$ higher 
($71.1\%$)
than
previous methods
on the challenging
Occlusion LINEMOD 
dataset,
and on average outperforming all other published results from the BOP benchmark for these 3 datasets.
Our code is available at \url{http://www.github.com/aaronwool/rcvpose}.
\end{abstract}

\section{Introduction}
\label{sec:Introduction}
Object pose estimation is an enabling technology for many applications including robot manipulation, human-robot interaction, augmented reality, and autonomous driving~\cite{pvnet,pavlakos20176,tremblay2018deep}. It is  challenging due to background clutter, occlusions, sensor noise, varying lighting conditions, and object symmetries. 
\begin{figure}[t]
\begin{center}
\includegraphics[width=0.7\columnwidth]{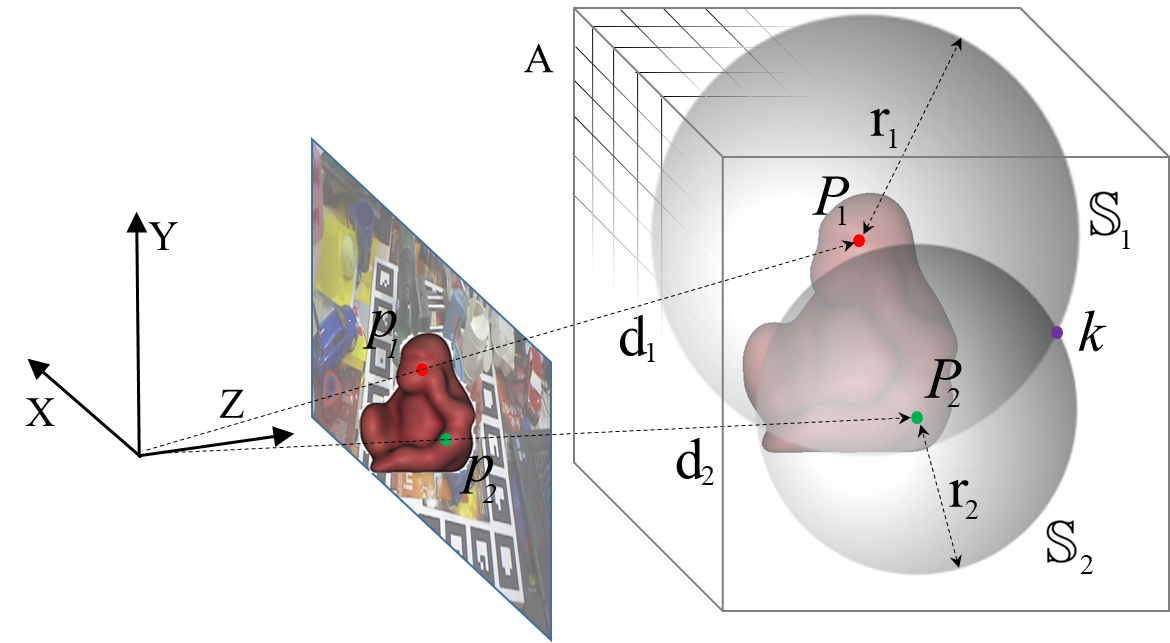}
  \caption{
  Radial voting scheme: 3D scene point $P_i$ at  depth d$_i$ projects to 2D image pixel $p_i$. The network estimates radial distance r$_i$ from $p_i$. 
  Sphere ${\mathbb{S}}_i$ 
  is centered at $P_i$ 
  with radius r$_i$,
  and
  all accumulator space A
  voxels
  on the surface of
  ${\mathbb{S}}_i$
  are incremented.
  Keypoint $k$ lies at the intersection
  of $\mathbb{S}_1
  \cap
  \mathbb{S}_2$, and all 
  other ${\mathbb{S}}_i$}

  \label{fig:radialVoting}
\end{center}
\end{figure}
Traditional methods have tackled the problem 
by establishing correspondences between a known 3D model and image features~\cite{hinterstoisser2012model,rothganger20063d}. They have generally relied on hand-crafted features and therefore fail when objects are featureless or when scenes are very cluttered and occluded~\cite{hu2019segmentation,pvnet}. Recent methods use deep learning and train end-to-end networks to directly regress an input image to a 6 DoF pose~\cite{kehl2017ssd,posecnn}. 
For example, CNN-based techniques have been proposed which regress 2D keypoints and use Perspective-n-Point (PnP) to estimate the 6 DoF pose parameters~\cite{pavlakos20176,yolo6d}. As an alternate to directly regressing keypoint coordinates, methods which \textit{vote} for keypoints have been shown to be highly effective~\cite{pvnet,posecnn,hu2019segmentation,houghnet}, especially when objects are partially occluded.
These schemes regress a distinct geometric quantity that relates positions of 2D pixels to 3D keypoints, and for each pixel casts this quantity into an accumulator space. As votes accumulate independently per pixel, these methods perform especially well in challenging occluded scenes.

While recent voting methods have shown great promise and leading performance, 
they
require the regression of either a 2-channel (for 2D voting)~\cite{pvnet} or 3-channel (for 3D voting )~\cite{PVN3d} activation map where voting quantities are accumulated in order to vote for keypoints.
The activation map is the image shaped
tensor where voting quantities are saved.
The dimensionality of the activation map follows from the formulation of the geometric
quantity being regressed,
and the
estimation errors in each channel tend to compound.
This leads to reduced localization accuracy
for higher dimensional activation maps when voting for keypoints. This 
observation has motivated our novel radial voting scheme, which regresses a one dimensional activation map for RGB-D data, leading to more accurate localization. The increase in keypoint localization accuracy also allows us to disperse our keypoint set farther, which increases the accuracy of transformation estimation, and ultimately that of 6 DoF pose estimation.


Our proposed method, \emph{RCVPose}, trains a
CNN to estimate the distance between a 3D 
keypoint, and the 3D scene point corresponding to each 2D RGB pixel. At inference, this distance is estimated for each 2D scene pixel, which is a 1D quantity and therefore
has the potential to be more accurate
than higher-dimension quantities regressed 
in previous methods.
For each pixel, a sphere of radius equal to this regressed distance is centered at each corresponding 3D scene point.
Those 3D accumulator space cells (\emph{voxels})
that intersect with the surface 
of these spheres are incremented,
and peaks indicate keypoint locations, as illustrated in Fig.~\ref{fig:radialVoting}. Executing this for minimally 3 keypoints allows the unique recovery of the 6 DoF object pose.

Our main contribution 
is a novel \textit{radial voting scheme} (based on a 1D regression)
which we experimentally show to be more accurate than previous voting schemes
(which are based on 2D and 3D regressions). 
Based on our radial voting scheme, 
a further contribution is a novel 6 DoF pose estimation method, called RCVPose. Notably, RCVPose requires only 3 keypoints per object, which is fewer than existing methods that use 4 or more keypoints~\cite{pvnet,PVN3d,houghnet}.
We experimentally characterize the performance of RCVPose on 3 standard datasets, and show that it outperforms previous peer-reviewed methods, performing especially well in highly occluded scenes. We also conduct experiments to justify certain design decisions and hyperparameter settings. 

\section{Related Work}
\label{sec:Related Work}
Estimating 6 DoF pose has been extensively addressed in the literature~\cite{lowe1999object,hinterstoisser2012model,posecnn,brachmann2014learning}. Recent deep learning-based methods use CNNs to generate  pose and can be generally classified into the three categories of  \emph{viewpoint-based}~\cite{hinterstoisser2012model}, 
\emph{keypoint-based}~\cite{posecnn},
and 
\emph{voting-based} methods~\cite{houghnet}.

\textbf{Viewpoint-based methods} 
predict 6 DoF poses by matching 3D or projected 2D templates.
In~\cite{park2019pix2pose}, a generative auto-encoder architecture 
used a GAN to convert RGB images into 3D coordinates, similar to the image-to-image translation task. Generated pixel-wise predictions were used in multiple stages to form 2D to 3D correspondences to estimate poses with RANSAC-based PnP. Manhardt~\etal~\cite{Manhardt_2019_ICCV}
proposed predicting several 6 DoF poses for each object instance to estimate the pose distribution generated by symmetries and repetitive textures. Each predicted hypothesis corresponded to a single 3D translation and rotation,
and estimated hypotheses collapsed onto the same valid pose when the object appearance was unique.
Recent variations include
Trabelsi~\etal~\cite{trabelsi2021ppn},
who used a
multi-task CNN-based encoder/multi-decoder network,
and
Wang~\etal~\cite{wang2020self6d}
and~\cite{labbe2020cosypose,Park2020NOL,Shao_2020_PFRL}, 
who used a rendering method by a self-supervised model on unannotated real RGB-D data to find an
optimal alignment.

\textbf{Keypoint-based methods} detect specified object-centric keypoints and apply PnP for final pose estimation. Hu~\etal~\cite{hu2019segmentation} proposed a segmentation-driven 6 DoF pose estimation method which used the visible parts of objects for local pose prediction from 2D keypoint locations. They then used the output confidence scores of a YOLO-based~\cite{redmon2018yolov3} network to establish 2D to 3D correspondences between the image and the object’s 3D model. Zakharov~\etal~\cite{zakharov2019dpod} proposed a dense pose object detector to estimate dense 2D to 3D correspondence maps between an input image and available 3D models,
recovering 6 DoF pose using PnP and RANSAC. In addition to RGB data, depth information was used in~\cite{PVN3d} to detect 3D keypoints with a Deep Hough Voting network,
with the 6 DoF pose parameters then fit with a least-squares method.


\textbf{Voting-based methods}
have a long history in pose estimation. 
Before 
artificial intelligence became widespread,
first 
the Hough Transform~\cite{duda1972hough}
and
RANSAC~\cite{fischler1981ransac} and
subsequently
methods such as pose clustering~\cite{Olson97efficientpose}
, image retrieval~\cite{brogan2021fast,schonberger2016vote} and geometric hashing~\cite{geometric-hashing}
were widely used to localize
 simple geometric shapes, objects in images
and full 6 DoF object pose. 
Hough Forests~\cite{HoughForest},
while learning-based,
still required hand-crafted feature descriptors. 
Voting was also extended to 3D point cloud images, 
such as 4PCS~\cite{amo_fpcs_sig_08} and 
its variations~\cite{4pcs,Super4PCS},
to estimate 
affine-invariant poses.

Following the advent of CNNs, hybrid methods emerged 
combining
aspects of both data-driven and classical voting approaches. 
Both
\cite{hu2019segmentation}
and 
\cite{pvnet}
conclude with RANSAC-based keypoint voting,
whereas
Deep Hough Voting~\cite{houghnet} proposed a complete MLP pipeline of keypoint localization using a series of convolutional layers as the voting module. 
To estimate keypoints, 
two different deep learning-based 
voting schemes 
have appeared~\cite{pvnet,posecnn,hu2019segmentation,houghnet}, the proposed  scheme introducing a third.
At training, all voting schemes regress a distinct quantity
that relates
positions of 
pixels to keypoints.
At inference, this quantity is estimated
for each pixel, and is cast into an
accumulator space in a voting process. 
Accumulator spaces
can cover the 2D~\cite{posecnn,hu2019segmentation,houghnet} 
image space,
or more recently 
the
3D~\cite{pvnet}
camera reference frame.
After voting, peaks in accumulator space indicate
 positions of  keypoints in the 
2D image or 3D camera frame.

While only a few hybrid voting-based methods exist for 6 DoF pose estimation,
they have outstanding performance,
which has motivated us to 
develop RCVPose as a further advance of this class of hybrid method.
Specifically, our method 
is inspired by PVNet~\cite{pvnet},
and is most closely related to 
the recently proposed 
PVN3D of
He~\etal~\cite{PVN3d},
which combined 
PVNet and Deep Hough Voting~\cite{houghnet} with a
3D accumulator space,
 utilizing the offset voting scheme of~\cite{posecnn}.

\section{Methodology}

\begin{figure}[t]
\centering
\hfill
\begin{subfigure}[b]{0.45\textwidth}
    \centering
    \includegraphics[width=0.7\textwidth]{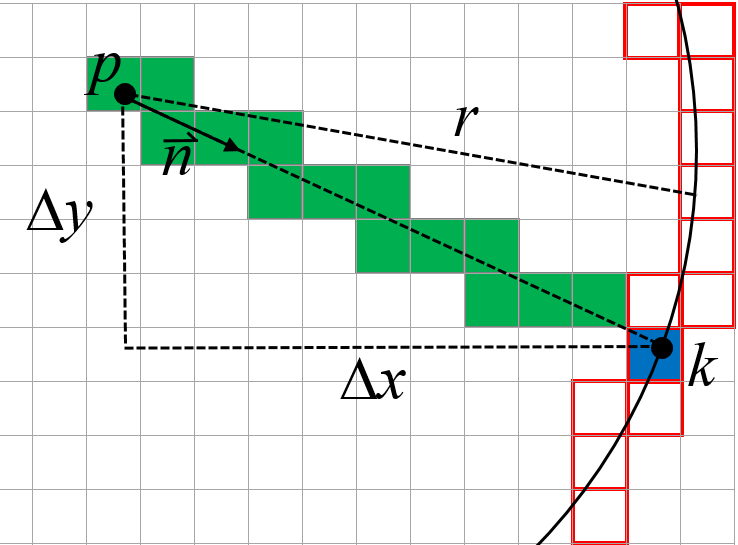} 
    \label{fig:2D voting}
    \caption{
    Votes cast (in 2D) for offset, vector, and radial voting}
\end{subfigure}
\hfill
\begin{subfigure}[b]{0.45\textwidth}
    \centering
    \includegraphics[width=0.7\textwidth]{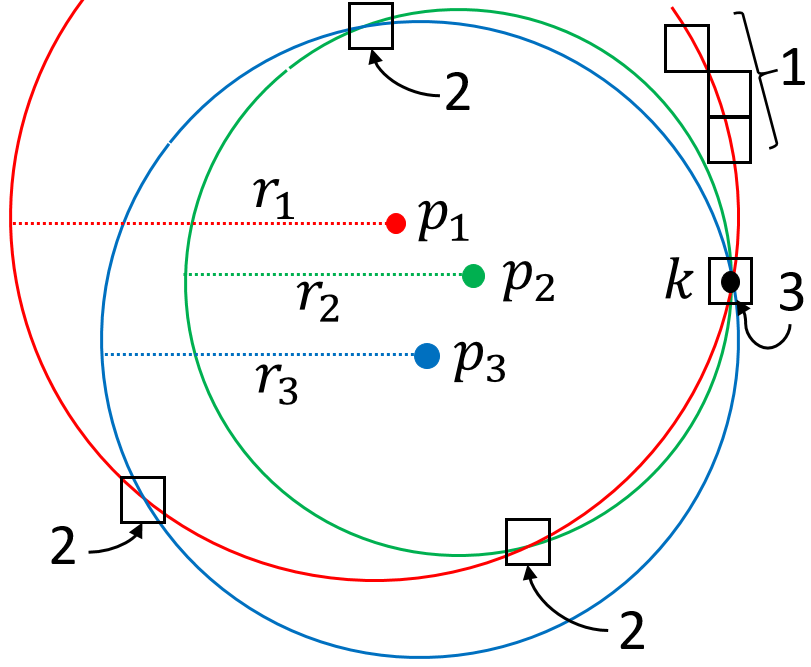}
    \label{fig:2D radial voting}
    \caption{
    Accumulator space values after radial voting for 3 points}
\end{subfigure}
  \caption{
  Keypoint Voting Schemes in 2D: a) Pixel $p$ casts votes for keypoint $k$ at blue bin (offset and vector voting), green bins (vector voting), and red bins (radial voting). b) Radial votes cast for pixels $p_1$, $p_2$, and $p_3$ result in bin peaks at the intersection of the circles, with the peak occuring at keypoint $k$ }
  \label{fig:allVoting}
\end{figure}

\subsection{Keypoint Voting Scheme Alternatives}
The three keypoint voting schemes are illustrated in 2D in Fig.~\ref{fig:allVoting}a, for image pixel $p$  and keypoint $k$ to be estimated. The grid represents the (initially empty) accumulator space bins, which are the voxel space elements where
votes are cast. In 
\emph{offset} voting,
the values of 
$\Delta x$ and $\Delta y$
are estimated from forward inference through the network.
These values are used to offset $p$ to reference that accumulator bin
(shown in blue)
containing $k$,
the value of which is then incremented.
Alternately, in 
\emph{vector} voting,
the direction
$\vec{n}$
is estimated,
and all bins (shown in green and blue) that intersect with $\vec{n}$ are incremented.
Finally, in 
\emph{radial} voting,
the scalar $r$ is estimated,
and all bins (shown outlined in red) are incremented that intersect with 
the perimeter of the circle of radius $r$ centered at $p$.
When repeated for all image pixels, the bin containing $k$ will contain the maximum accumulator space value, irrespective of which scheme is used, so long as the quantities estimated by network inference are sufficiently accurate.
In Fig.~\ref{fig:allVoting}b,
circles generated by
radial voting are illustrated for three image pixels.
Each bin contains a count of the number of circle perimeters that it intersects, such that the peak value of 3 indicates the location of keypoint $k$.
The above three voting schemes extend directly to 3D space,
in which the accumulator space is a grid of voxels,
the offset scheme contains an additional $\Delta z$ component,
$\vec{n}$ is a 3-dimensional vector, and the radial scheme casts votes on the surfaces of 3D spheres rather than 2D circles.

Formally, let 
$p_i$ be pixel
from RGB-D image $I$
with 
2D image coordinate
$(u_i,v_i)$
and corresponding
3D camera frame coordinate
$(x_i,y_i,z_i)$.
Further 
let 
$k^{\theta}_j\!=\!(x_j,y_j,z_j)$
denote 
the camera frame coordinate of the $j^{th}$ keypoint of
an object
located at
6 DoF pose 
$\theta$. 
The quantity ${\bf m_o}$ regressed in the first \textit{offset} scheme~\cite{hu2019segmentation,houghnet}
is the displacement between the two 3D points, 
denoted as
$
    {\bf m_o}\!=\!(\Delta x, \Delta y, \Delta z)\!=\!(x_i\!-\!x_j,\; y_i\!-\!y_j,\; z_i\!-\!z_j)
$.
Alternately, the 3D quantity 
 ${\bf m_v}$ 
from the second 
\textit{vector}
scheme~\cite{pvnet,posecnn} is the unit vector
pointing to 
$k^{\theta}_j$
from 
$p_i$, denoted as
$
    {\bf m_v}\!=\!(dx, dy, dz) 
\!=\!\frac{\bf m_o}{\lVert {\bf m_o} \rVert} 
$.
The 3D vector scheme can alternately be
parametrized into a 2D
\emph{polar}
scheme,
denoted as
$
{\bf m_p} 
\!=\!(\phi,\psi)
\!=\!(\cos^{-1}{dz},
\tan^{-1}\frac{dy}{dx})
$.
Finally, the 1D quantity  ${\bf m_r}$ from the 
 \textit{radial} scheme 
 proposed here
 is simply the Euclidean distance between the points, i.e.
$
    {\bf m_r}\!=\!\lVert{\bf m_o}\rVert
$.

The above quantities encode different
information
about the relationship
between $p_i$
and 
$k^{\theta}_j$.
For example,
${\bf m_v}$,
${\bf m_p}$,
and 
${\bf m_r}$
can be derived directly from
${\bf m_o}$,
whereas 
${\bf m_o}$
cannot be derived from the others.
Also, 
${\bf m_r}$
and
${\bf m_v}$ 
(and ${\bf m_p}$)
are independent of one another.
This difference in geometric
information leads to their different
 dimensionality, and ultimately the greater
accuracy of radial voting,
as discussed in Sec.~\ref{sec:Voting Scheme Accuracy}.

\begin{figure*}[t]
\begin{center}
\includegraphics[width=0.85\textwidth]{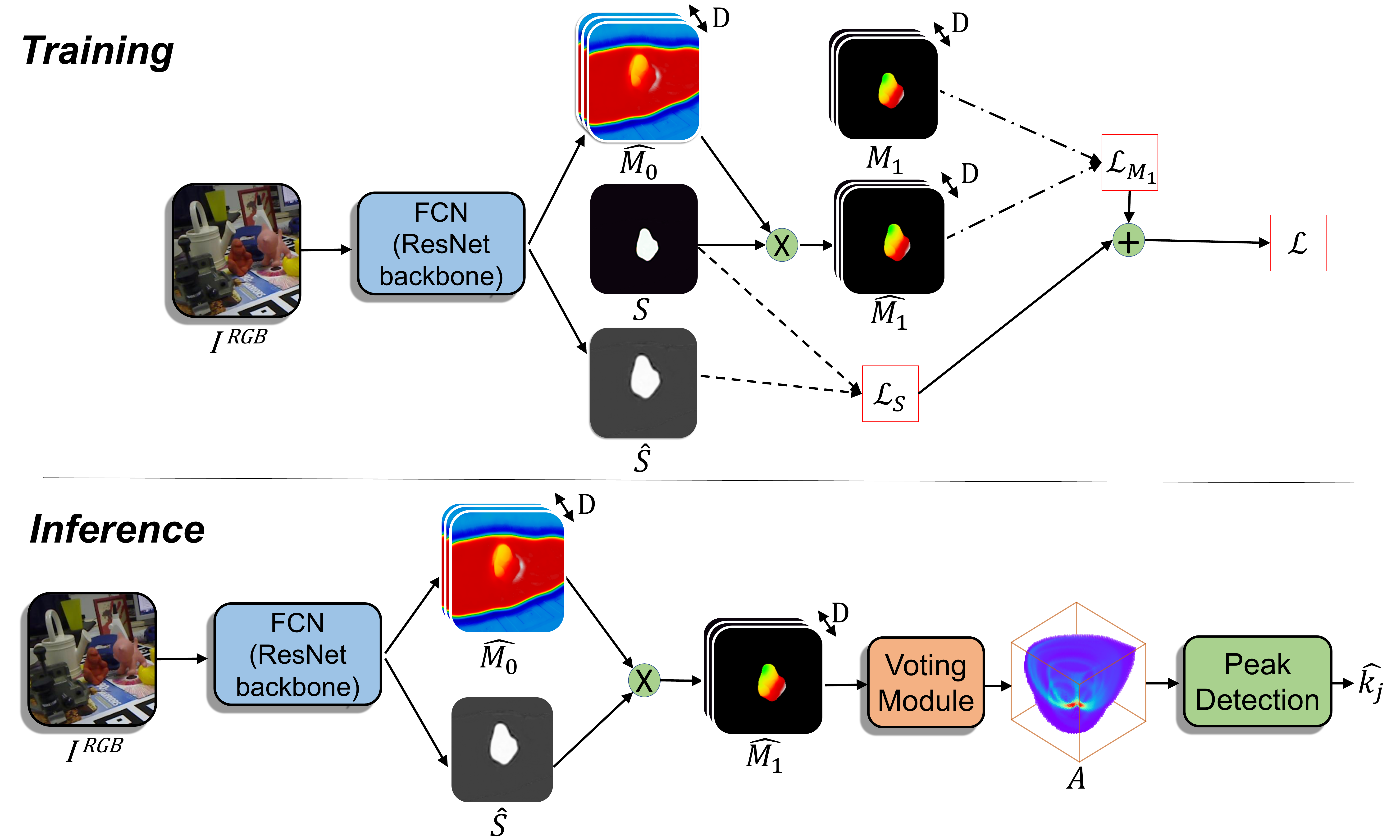}
\caption{
RCVPose training and inference. $\widehat{M}_0$,
  $\widehat{M}_1$,
  and $M_1$
  have channel depth $D\!=\!1$
  for radial,
  $D\!=\!2$ for polar,
  or $D\!=\!3$ for offset or vector voting schemes\label{fig:train&test}}
  
\end{center}
\end{figure*}

\subsection{Keypoint Estimation Pipeline}
\label{sec:architecture}
The above described voting schemes can be used
interchangeably within a keypoint estimation pipeline.
The training inputs
(Fig.~\ref{fig:train&test})
are:
RGB fields $I^{RGB}$ of image $I$;
ground truth binary segmented image $S$ of the foreground object
at pose $\theta$; ground truth keypoint coordinate $k^{\theta}_j$, and;
the ground truth voting scheme
values 
(i.e. one of
${\bf m_o}$,
${\bf m_v}$,
${\bf m_p}$
or 
${\bf m_r}$)
for each pixel in $S$, represented by matrix $M_1$.
$M_1$ is calculated for a given 
$k^{\theta}_j$
using one of
the voting scheme values,
and
has either channel depth $D\!=\!3$ 
for 
${\bf m_o}$
or  
${\bf m_v}$,
$D\!=\!2$
for 
${\bf m_p}$,
or $D\!=\!1$
for  
${\bf m_r}$.
Both $S$ and $M_1$
are assessed to compute the loss
$\mathcal{L}$ as:
\begin{equation}
\mathcal{L} = \mathcal{L}_{S}+\mathcal{L}_{M_1} ,
\label{eq:Loss}
\end{equation}
\begin{equation}
\mathcal{L}_{S} = \frac{1}{N}\sum_{i=1}^{N} \left |\widehat{S}_i - S_i  \right | ,
\label{eq:LossS}
\end{equation}
\begin{equation}
\mathcal{L}_{M_1} = \frac{1}{N}\sum_{i=1}^{N} 
\left( |\widehat{M_1}_i - {M_1}_i  | \right) ,
\label{eq:LossM1}
\end{equation}
with summations 
 over all $N$ pixels.
The network output is
 estimate $\widehat{S}$ of $S$,
and  (unsegmented) estimate $\widehat{M}_0$
of $M_1$. 

At inference (Fig.~\ref{fig:train&test}),
$I^{RGB}$ is fed to the network
which returns estimates
$\widehat{S}$ 
and $\widehat{M_0}$,
the
element-wise multiplication of which
yields segmented estimate
$\widehat{M}_1$.
Each pixel $(u_i,v_i)$ of $\widehat{M}_1$,
with corresponding 3D coordinate
$(x_i,y_i,z_i)$
drawn from the depth field $I^D$ of $I$,
then independently casts a vote 
through the voting module
into 
the initially empty 3D accumulator space
$A$. 

Vote casting is performed
for each 
$(u_i,v_i)$,
and
is distinct for each voting scheme. In 
\emph{offset} voting,
accumulator space $A$ bin
$A[
x_i\!+\!\widehat{M}_1[u_i,v_i,0], \ 
y_i\!+\!\widehat{M}_1[u_i,v_i,1],
z_i\!+\!\widehat{M}_1[u_i,v_i,2]
]
$
is incremented, thereby voting for the specific bin
of $A$ that contains keypoint $k^{\theta}_j$.
In
\emph{vector} 
and 
\emph{polar}
voting,
every $A$ bin is incremented that intersects with the
ray 
$\alpha 
(
x_i\!+\!\widehat{M}_1[u_i,v_i,0],\ 
y_i\!+\!\widehat{M}_1[u_i,v_i,1],\ 
z_i\!+\!\widehat{M}_1[u_i,v_i,2]
)$, 
for 
$\alpha > 0$,
thereby casting a vote for every bin along the ray
that intersects with
$(x_i,y_i,z_i)$ and $k^{\theta}_j$.
Finally, in
\emph{radial} voting,
every $A$ bin is incremented that intersects with the
sphere 
of radius $\widehat{M}_1[u_i,v_i]$
centered at 
$(x_i,y_i,z_i)$,
thereby voting for 
every bin that 
lies on the surface of a sphere 
upon which
$k^{\theta}_j$
resides.
Whichever scheme is used, at the  conclusion of vote casting for all 
$(u_i,v_i)$,
a global peak will exist in the
$A$ bin containing $k^{\theta}_j$, and  a simple peak detection
operation is then sufficient to estimate keypoint position
$\widehat{k}^{\theta}_j$, within the precision of $A$.
The radial voting scheme has 
been shown to be more accurate 
than the other schemes at keypoint estimation,
as shown
in the experiments in 
Sec.~\ref{sec:Voting Scheme Accuracy}.

\subsection{RCVPose}
\label{sec:Overview}
The above keypoint voting method  formed the core of RCVPose. 
Radial voting was used, based on its superior accuracy
as demonstrated in
Sec.~\ref{sec:Voting Scheme Accuracy}.
The network of Fig.~\ref{fig:train&test} was used with ResNet-152 
as the FCN-ResNet module.
The minimal $K\!\!=\!3$ keypoints were 
used for each object,
selected
from the corners of each object's bounding box. 
Based on
Sec.~\ref{sec:Keypoint Scaling},
keypoints were scaled to lie
beyond the surface
of each object,
$\sim 2$ object radius units from its centroid.

The network structure was
based on 
a Fully Convolutional ResNet-152~\cite{2015Resnet},
similar to PVNet~\cite{pvnet}, 
albeit with 
two main differences.
First, we replaced LeakyReLU with 
ReLU as the activation function.
This was because
our radial voting scheme only
includes positive values,
in contrast to the vector voting scheme of PVNet
which
also admits negative values.
Second, 
we increased the number of skip connections
linking the downsampling and 
upsampling layers
from three to five,
to include extra local features when
upsampling~\cite{2015fcn}.

All voxels were initialized to zero,
with their values incremented
as votes were cast.
The voting process is
similar to 3D sphere rendering,
wherein those voxels that intersect with 
the sphere surface have their values incremented.
The process is based on Andre's circle
rendering algorithm~\cite{andres1994discrete}.
We generate a series of 2D slices of $A$ 
parallel to the x-y plane, that fall within the sphere radius 
from the sphere center
in 
both directions of the
z-axis.
For each slice, the radius of the circle
formed by the intersection of the sphere and
that slice is calculated,
and all voxels that intersect with this circumference
are incremented.
The algorithm is accurate and efficient,
requiring that only a small portion of the voxels
be visited for each sphere rendering.
%
It was implemented in Python
and parallelized at the thread level,
and executes with an efficiency
similar to forward network inference.

Once the $K\!\!=\!3$ keypoint locations
are estimated for an image,
it is straightforward to 
determine the object's 6 DoF rigid transformation 
$\theta$,
from the corresponding estimated scene and 
ground truth object 
keypoint coordinates~\cite{Horn:88,Egg+Fis:98}.
This is analogous to the approach of~\cite{PVN3d},
and 
is efficient compared to previous pure RGB approaches~\cite{pvnet} which
employ an 
iterative 
PnP method.

\section{Experiments}
\label{sec:Experiments}

\subsection{Datasets}
\label{sec:Datasets}
The 
\textbf{LINEMOD} dataset~\cite{hinterstoisser2012model} includes $1200$ images per object. The training set contains only 180 training samples using the standard $15\%/85\%$ training/testing split~\cite{posecnn,pvnet,bukschat2020efficientpose,PVN3d,hu2019segmentation}. 
We augmented the dataset by 
rendering the objects with a random rotation and translation,
transposed 
using the BOP rendering kit~\cite{hodavn2020bop}
onto a background
image drawn from the MSCOCO dataset~\cite{lin2014coco}.
An additional 1300 augmented images
were generated for each object in this way,
 inflating the training set to 1480 images per object.

The LINEMOD depth images have an 
offset compared to the ground-truth pose values, for unknown reasons~\cite{manhardt2018LMOffset}. 
To reduce the impact of this offset, we regenerated 
the depth field for each training image 
from the ground truth pose,
by reprojecting the depth value drawn from the
object pose at each 2D pixel coordinate.
The majority (1300) of the 
resulting training
set were in this way purely synthetic images,
and the minority (180) comprised real RGB and synthetic depth.
All test images were original, real and unaltered.
%

\textbf{Occlusion LINEMOD}~\cite{brachmann2014learning} is a re-annotation of LINEMOD comprising a subset of 1215 challenging test images
of 
partially occluded objects. 
The protocol is to train using LINEMOD images only,
and then test on Occlusion LINEMOD to verify robustness. 

\textbf{YCB-Video}~\cite{posecnn} 
is a much larger dataset,
containing
130K key frames of
21 objects over 92 videos.
We split 113K frames for training and 27K frames for testing,  
following PVN3D~\cite{PVN3d}. For data augmentation, YCB-Video provides 80K synthetic images with random object poses, rendered on a  black background. 
We repeated here the process described above, by rendering random MSCOCO images
as background. The complete training dataset therefore
comprised 113K real + 80K synthetic = 193K images.

\subsection{Implementation Details}
\label{sec:Implementation Details}

Prior to training,
each RGB image is shifted
and scaled
to adhere to the ImageNet mean
and standard deviation~\cite{Imagenet}.
The 3D coordinates were calculated
from the image depth fields
and represented in decimeter units,
as all LINEMOD 
and YCB-Video  objects
are at most 1.5 decimeters in diameter and the backbone network can estimate better when the output is within a normalized range.
%
The loss functions of Eqs.~\ref{eq:Loss}-\ref{eq:LossM1}
were used
with an Adam optimizer, 
with
initial learning rate lr=1e-4.
The lr was adjusted on a fixed schedule, 
re-scaled by a factor of $0.1$ every 70 epochs.
The network trained for 
300 and 500
epochs
for each object in the LINEMOD
and YCB-Video datasets respectively,
with 
batch size 32.

The accumulator space $A$ is represented as a
flat 3D integer array, i.e. an
axis-aligned grid of voxel cubes.
The size of $A$
was set 
for each test image 
to the bounding box
of the 3D data.
The voxel resolution
was set to 5 mm, 
which was found to be 
a good tradeoff between memory expense and keypoint localization accuracy (see
Supplementary Material Sec. S.4.5).

For each object, 
3 instances of the network were trained, 
one for each keypoint. We also implemented a version in which all 3 keypoints were trained simultaneously, within a single network. In this version, the $\hat{M}_0$,
$\hat{M}_1$, and $M_1$ representations of
Fig.~\ref{fig:train&test} are replicated 3 times, and the FCN-ResNet weights are shared.
Our experiments (detailed in the supplementary material) showed that the accuracy was poorer for this version, than when using separate networks for each keypoint. The only two methods that have used a combined network for all keypoints and all objects are
GDRNet~\cite{wang2021gdr} and SOPose~\cite{di2021so},
against which our performance compares favourably (see Sec.~\ref{sec:Results}).

\subsection{Evaluation Metrics}
\label{sec:Evaluation Metrics}
We follow the ADD(s) metric 
defined by~\cite{hinterstoisser2012model} 
to evaluate LINEMOD,
whereas YCB-Video is evaluated based on both ADD(s) and AUC as proposed by~\cite{posecnn}. 
All metrics are based on
the distances between
corresponding points 
as objects are transformed
by the ground truth and estimated transformations.
ADD measures the average distance between
corresponding points,
whereas ADDs averages the minimum distance between
closest points, and is more forgiving for symmetric
objects.
A pose is considered correct if its ADD(s) falls within 10$\%$
of the object radius.
AUC applies the ADD(s) values
to determine the success of
an estimated transformation,
integrating these results 
over a varying 0 to 100 mm threshold.

\subsection{
Comparison of Keypoint 
Voting Schemes
}
\label{sec:Voting Scheme Accuracy}

We first conducted an experiment to evaluate the
relative accuracies of the four voting schemes
at keypoint localization,
using the process from Sec.~\ref{sec:architecture}.
Each scheme used the same
$15\%/85\%$ train/test split of 
a subset of objects from the LINEMOD dataset. All four schemes used the exact same backbone network and hyperparameters.
Specifically,
they all 
used a fully convolutional ResNet-18~\cite{2015fcn},
batch size 48, initial learning rate 1e-3, 
and Adam
optimizer,
with accumulator space resolution 
of 1 mm.
They were all trained with a fixed learning rate reduction schedule,
which reduced the rate by a factor of 10 following every 70 epochs,
and all trials trained
until they fully converged.

\begin{table}[t]
\begin{center}
\caption{
Keypoint localization error $\bar{\epsilon}$, 
for surface (FPS) and disperse keypoints:  
mean 
$\mu$
and standard deviation 
$\sigma$
for 4 voting schemes
$\{v,o,p,r\}$,
with
$\bar{r}$
= mean keypoint distance to object centroid
\label{tab:KEE-FPS-BBox-2}}
\begin{tabular}{r|c|c|cc|cc|cc|cc}
\multicolumn{3}{c}{} &
\multicolumn{8}{c}{$\bar{\epsilon}$ [mm]} \\
\cline{4-11}
\multicolumn{3}{l}{} & \multicolumn{2}{c}{vector (3D)}    & \multicolumn{2}{c}{offset (3D)}                & \multicolumn{2}{c}{polar (2D)}                    & \multicolumn{2}{c}{ radial (1D)}               \\[-5pt]
 \multicolumn{2}{c}{}
 &
 \multicolumn{1}{c|}{$\bar{r}$ [mm]}
&
\multicolumn{1}{c}{$\mu_{v}$} & \multicolumn{1}{c}{$\sigma_{v}$} & \multicolumn{1}{|c}{$\mu_{o}$} & \multicolumn{1}{c|}{$\sigma_{o}$} & \multicolumn{1}{|c}{$\mu_{p}$} & \multicolumn{1}{c|}{$\sigma_{p}$} & \multicolumn{1}{c}{$\mu_r$} & \multicolumn{1}{c}{$\sigma_r$} \\ \hline \hline
ape & 
 \parbox[t]{3mm}{\multirow{3}{*}{\rotatebox[origin=c]{90}{FPS}}} 
& 61.2 & 10.0 & 5.8 & 5.8 & 2.6 & 5.6 & 2.4 & {\bf 1.3} & {\bf 0.7} \\ 
driller & & 129.4 & 10.0 & 2.3 & 6.5 & 4.7 & 5.3 & 2.5 & {\bf 2.2} & {\bf 1.0} \\ 
eggbox & & 82.5 & 11.8 & 5.3& 5.2 & 2.7 & 4.9 & 1.9 & {\bf 2.0} & {\bf 0.7}\\ 
 \hline

 ape & 
 \parbox[t]{3mm}{\multirow{3}{*}{\rotatebox[origin=c]{90}{disperse}}} 
 & 142.1 & 12.5 & 7.6 & 10.4 & 5.3 & 5.7 & 2.5 & {\bf 1.8} & {\bf 0.8} \\ 
driller & & 318.8 & 11.3 &  8.2 & 9.5 & 3.5 &  5.2 &  2.6 &  {\bf 2.7} &  {\bf 0.8} \\ 
eggbox & & 197.3 & 13.7 &  8.5 & 11.4 & 4.7 &  7.2 &  3.4 &  {\bf 2.4} &  {\bf 1.2} \\ 
 \hline
\end{tabular}
\end{center}
\label{tab:KEE-FPS-BBox}
\end{table}

The only difference between
trials,
other than the selective use of
either
${\bf m_o, m_v, m_p}$ 
or 
${\bf m_r}$
in training
 $\widehat{M}_1$,
was a slight variation in the loss functions. For 
${\bf m_o}$ 
and 
${\bf m_r}$, 
the L1 loss from Eqs.~\ref{eq:Loss}-\ref{eq:LossM1}
was used,
identical to the offset voting
in PVN3D~\cite{PVN3d}.
Alternately, 
for
${\bf m_v}$
and 
${\bf m_p}$, 
the Smooth L1 equivalents
of 
Eqs.~\ref{eq:LossS} and \ref{eq:LossM1}
(with $\beta\!\!=\!\!1$)
were used, 
as in PVNet~\cite{pvnet}
(albeit therein using a 2D accumulator space).

\subsubsection*{Surface Keypoints:}
Sets
of size 
$K$=\;4
surface keypoints
were selected for each object tested,
using the Farthest Point Sampling
(\emph{FPS})
method~\cite{eldar1997farthest}. 
FPS selects 
points on the surface of an object which are well separated,
and is a popular keypoint generation strategy~\cite{pvnet,PVN3d,qi2017pointnet++,houghnet}. 
Following training, 
each keypoint's location
$\widehat{k}_j^{\theta_i}$
was estimated
by 
passing
each test image 
$I_i$ through the network, as in Fig.~\ref{fig:train&test}.
The error 
$\epsilon_{i,j}$
for each estimate 
was 
its Euclidean distance from its
ground truth location, i.e.
$\epsilon_{i,j}\!=\!\lVert\widehat{k}_j^{\theta_i}\!-\!k_j^{\theta_i}\rVert$.
The average of
$\epsilon_{i,j}$
for an object over all test images and keypoints was the \emph{keypoint estimation error}, denoted as 
$\bar{\epsilon}$.

Each 
voting scheme was implemented with care,
so that they were numerically accurate and equivalent.
To test the correctness of voting in isolation,
ground truth
values of $M_1$
calculated for each object and voting scheme
were passed directly into the voting module, effectively replacing 
$\widehat{M}_1$ with $M_1$ in 
the inference stage of Fig.~\ref{fig:train&test}.
For each voting scheme,
the average
$\bar{\epsilon}$
for all objects was
similar and less than the accumulator
space resolution of 1 mm,
indicating that the 
implementations were correct and accurate.

The
$\bar{\epsilon}$ values
were evaluated
for the four voting schemes
for the ape, driller and eggbox 
LINEMOD
objects
as summarized in 
Table~\ref{tab:KEE-FPS-BBox-2}.
These three particular objects
were chosen as the ape is the
smallest
and the driller the largest
of the objects,
whereas the eggbox includes a rotational
symmetry. 
Table~\ref{tab:KEE-FPS-BBox-2}
includes a measure of
the average distance 
$\bar{r}$
of
the ground truth keypoints to each object centroid.
Radial voting 
is seen to be the most accurate method,
with a mean value
1.9-4.3x more accurate than 
the next most accurate polar voting, with smaller standard deviations. 
Notably, the ordinal relationship between the four schemes
remains consistent across the 
scheme dimensionality,
which indicates that dimensionality
impacts keypoint localization error.

\subsubsection*{Disperse Keypoints:} 
We repeated this experiment
for keypoints  
selected from the corners of each object's bounding box, which was first scaled by a factor of 2 so that
the keypoints
were dispersed to
fall outside of the object's surface.
The results in
Table~\ref{tab:KEE-FPS-BBox-2} indicate that radial voting still outperforms the other
two schemes by a large margin.
Whereas the other
two methods decrease in accuracy sharply as the mean keypoint distance 
$\bar{r}$ 
increases,
radial voting accuracy degrades more gracefully. For example,
for the ape, the 
232$\%$
increase in $\bar{r}$ 
from 61.2 to 142.1 mm,
reduced
accuracy for 
offset voting
by $80\%$ 
(from 5.8 to 10.4 mm),
but only by 
$23\%$ 
(from 2.2 to 2.7 mm)
for radial voting.

The improved accuracy of radial
voting is likely due to the fact that the radial scheme regresses a 1D quantity,
compared with 
the 2D polar, and the 3D
offset and vector scheme quantities.
It seems likely that the errors in each independent dimension compound during voting. 
This is further supported by the recognition that the
polar scheme
is simply a reduced dimensionality
parametrization of the vector scheme,
and yet its performance is far superior, with between 1.7-2.4x greater accuracy.
Radial voting also has a degree of resilience to rotations,
which is lacking in 
the other schemes.
Specifically,
the three voting quantities
${\bf m_o}$,
${\bf m_v}$,
and
${\bf m_p}$
are all sensitive to object in-plane rotations,
whereas only radius scheme
${\bf m_r}$
is invariant to in-plane rotations.

\subsection{Keypoint Dispersion}
\label{sec:Keypoint Scaling}
%
%
\label{sec:Keypoint Dispersion}

\subsubsection{Impact on Transformation Estimation:}
\label{sec:impact1}
It was suggested in~\cite{pvnet}
that 6 DoF pose estimation accuracy
is improved by 
selecting  keypoints
that lie on the object
surface,
rather than the bounding box corners which lie just
beyond the object surface.
This may be  
the case when keypoint localization error increases signficantly with keypoint disperson, as 
occurs with vector and offset voting. 
There is, however,
an advantage to 
dispersing the
keypoints farther apart
when using radial voting,
which has a
lower 
estimation error.

To demonstrate this, we
conducted an experiment in which the keypoint locations were dispersed
to varying degrees under a constant keypoint estimation error, with the impact measured on the accuracy of the resulting estimated transformation. We first selected
a set $\cal{K}$=$\{k_j\}_{j=1}^{4}$  keypoints on the surface of an object, using the FPS strategy. This
set was then rigidly transformed by $T$, comprising a random
rotation (within $0^\circ$ to $360^\circ$ for each axis) and a random translation (within 1/2 of the object radius),
to form keypoint set ${\cal{K}}_T$.
Each keypoint in ${\cal{K}}_T$ was
then independently pertubed by a magnitude of 1.5 mm in a random direction, to simulate the keypoint estimation error of the radial voting scheme,
resulting in (estimated) keypoint set $\widetilde{\cal{K}}_{\widetilde{T}}$.

Next, the estimated transformation $\widetilde{T}$ between 
$\widetilde{\cal{K}}_{\widetilde{T}}$ and the original (ground truth) keypoint set $\cal{K}$
was calculated using the Horn method~\cite{Horn:88}. This process simulates the
pose estimation that would occur between estimated keypoint locations, each with some error, and their corresponding ground truth model keypoints. The surface points of the object were then transformed by both the ground truth $T$ 
and the estimated $\widetilde{T}$ transformations, and the  distances separating corresponding transformed surface points were compared, as a measure of the accuracy of the estimated transformation.

The above process was repeated for versions of $\cal{K}$
that were dispersed by scaling an integral factor of the object radius from the object centroid. The exact same error perturbations (i.e. magnitudes and directions) were applied to each keypoint for each new scale value. The scaled trials therefore represented keypoints that were dispersed more distant from the object centroid, albeit with the exact same localization error.

\begin{figure}[t]
\centering
\begin{subfigure}[b]{.5\columnwidth}
  \centering
  \includegraphics[width=0.9\columnwidth]{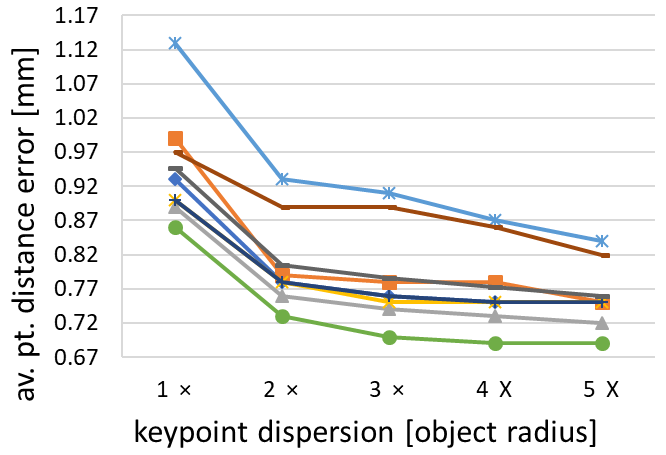}
  \caption{
  Transformation error ($\downarrow$)}
  \label{fig:dispersionPlot-a}
\end{subfigure}%
\begin{subfigure}[b]{.5\columnwidth}
    
  \centering
  \includegraphics[width=0.9\columnwidth]{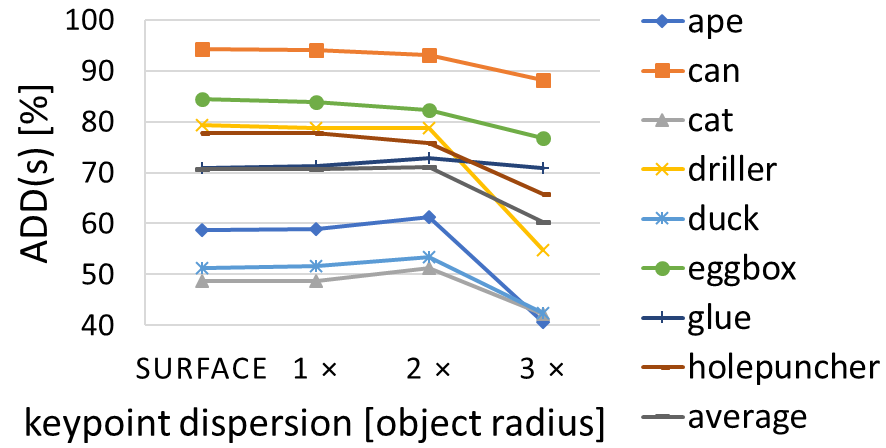}
  
  \caption{
  6 DoF pose ADD(s) ($\uparrow$)}
  \label{fig:dispersionPlot-b}
\end{subfigure}
\caption{
Impact of keypoint dispersion on (a) Transformation estimation error, and (b) 6 DoF pose estimation ADD(s)}
  \label{fig:dispersionPlot}
\end{figure}

This process was executed for all Occlusion LINEMOD objects, with 100 trials for each scale factor value from 1 to 5. The means 
of the corresponding point distances
(i.e. the ADD metric as defined in~\cite{hinterstoisser2012model})
are 
plotted in Fig.~\ref{fig:dispersionPlot-a}. 
It can be seen that ADD decreases for the first few scale factor increments for all objects,
indicating an improved transformation estimation accuracy for larger keypoint dispersions.
This increase in accuracy stems from improved rotational estimates, as the same positional perturbation error
of a keypoint
under a larger moment arm will result in a smaller angular error. The translational component of the
transformation is not impacted by the scaling, as the Horn method starts by centering the two point clouds. After a certain increase in scale factor of 3 or 4, the unaffected translational error dominates, and the error plateaus.

This experiment shows
that the transformation estimate
from corresponding 
ground truth and estimated keypoints will be more
accurate,
when the keypoints are 
dispersed further
($\sim\!\!1$ object radius, i.e. a scale of 2x) from the
object's surface, when keypoint estimation error
itself remains small
($\sim\!\!1.5$ cm).
\subsubsection{Impact on 6 DoF Pose Estimation:}
\label{sec:impact2}
The above result can be leveraged 
to further improve the accuracy of 6 DoF pose estimation
when using radial voting.
An experiment was executed
for all Occlusion LINEMOD objects for varying keypoint dispersions. The keypoints were first selected to lie on the
surface of each object using FPS,  
and the complete RCVPose inference pipeline was executed,
yielding an ADD(s) value for each trial image. The keypoints were then projected outward from each object's centroid to a distance of 1, 2 and 3 object radius values, and 
RCVPose inference was once again executed and ADD(s)  recalculated.

The results are plotted in Fig.~\ref{fig:dispersionPlot-b}. Of the 8 objects, 4 had a higher ADD(s) value at a dispersion of 2x, as did the average over all objects. 
It seems that the 
decreased 
transformation estimation error
(Fig.~\ref{fig:dispersionPlot-a}) at 2x radius dispersion
more than compensates for the
gradual increase in keypoint localization error exhibited by radial voting.
\begin{table}[t]
\begin{minipage}[t]{.5\linewidth}
\caption{
LINEMOD and Occlusion \\
LINEMOD accuracy results\label{tab:LINEMOD+OccLINEMOD}}
\centering
\begin{tabular}{llcc}
\cline{3-4}
\multicolumn{2}{c}{} &
\multicolumn{2}{c}{ADD(s) [$\%$]} \\
\hline
Mode      & Method    
& LM & O-LM
\\\hline 
&  SSD6D~\cite{kehl2017ssd} & 9.1 & - \\ 
& Oberweger~\cite{oberweger2018making}  & - & 27.1 \\
& Hu et al.~\cite{hu2019segmentation} & - & 30.4 \\
& Pix2Pose~\cite{park2019pix2pose}  & 72.4  & 32.0 \\
& DPOD~\cite{zakharov2019dpod}      & 83.0 & 32.8 \\
& PVNet~\cite{pvnet}      & 86.3 & 40.8 \\
\multirow{-7}{*}{RGB}
& DeepIM~\cite{li2018deepim} & 88.6 & - \\ 
& PPRN~\cite{trabelsi2021ppn}    & 93.9     & 58.4 \\
& GDR-Net~\cite{wang2021gdr}  & 93.7  & 62.2 \\
& SO-Pose~\cite{di2021so}  & 96.0  & 62.3 \\\hline
 & YOLO6D~\cite{yolo6d}      & 56.0 & 6.4  \\
 & SSD6D+ref~\cite{kehl2017ssd} & 34.1 & 27.5 \\
                              
\multirow{-3}{*}{RGB}      
&  PoseCNN~\cite{posecnn}      & - & 24.9 \\
\multirow{-3}{*}{+D ref}
& DPOD+ref~\cite{zakharov2019dpod}   & 95.2 & 47.3 \\\hline
& DenseFusion~\cite{wang2019densefusion}        & 94.3 &  - \\
& PVN3D~\cite{PVN3d}      & 99.4  & 63.2 \\
& PR-GCN~\cite{zhou2021pr}          & 99.6  & 65.0 \\
& FFB6D~\cite{he2021ffb6d}& {\bf 99.7}  & 66.2 \\
& RCVPose     & 99.4 & 70.2 \\
\multirow{-6}{*}{RGB-D}  
& RCVPose+ICP  & {\bf 99.7} & {\bf 71.1}\\ \hline
\end{tabular}
\end{minipage}
\begin{minipage}[t]{.5\linewidth}
\caption{
YCB-Video accuracy results
\label{tab:ycbvideo}
} 
\centering
\begin{tabular}{clcc}
\hline  
\begin{tabular}{c} 
D ref? 
\end{tabular} 
& 
\begin{tabular}{c} 
Method 
\end{tabular} 
& 
\begin{tabular}{c} 
\mbox{ADD(s)} 
\end{tabular} 
&
\begin{tabular}{c} 
\mbox{AUC} 
\end{tabular} 
\\ [3pt]
\hline
\multicolumn{1}{c}{{}}                          & {PoseCNN~\cite{posecnn} } 
& 59.9 &  75.8
\\ 
\multicolumn{1}{c}{{}}                              & {DF (per-pixel)~\cite{wang2019densefusion}}  
& 82.9 & 91.2
\\
\multicolumn{1}{c}{{}} & {SO-Pose~\cite{di2021so}}
& 56.8 & 90.9
\\ 
\multicolumn{1}{c}{{}} &
{GDR-Net~\cite{wang2021gdr}}
& 60.1 & 91.6
\\ 
\multicolumn{1}{c}{{}} & {PVN3D~\cite{PVN3d}}
& 91.8 & 95.5
\\ 
\multicolumn{1}{c}{{}} & {PR-GCN~\cite{zhou2021pr}}
& - & 95.8
\\
\multicolumn{1}{c}{}                          & {FFB6D~\cite{he2021ffb6d}} 
& 92.7 & \textbf{96.6}\\
 \multicolumn{1}{c}{\multirow{-7}{*}{No}}     & RCVPose                              
 & \textbf{95.2} & \textbf{96.6}
 \\ \hline
\multicolumn{1}{c}{}                          & PoseCNN~\cite{posecnn} 
& 85.4 & 93.0
\\ 
 \multicolumn{1}{c}{}                          & {DF (iterative)~\cite{wang2019densefusion}} 
 & 86.1 & 93.2
 \\ 
\multicolumn{1}{c}{}                          & {PVN3D~\cite{PVN3d}+ICP} 
& 92.3 & 96.1
\\ 
\multicolumn{1}{c}{}                          & {FFB6D~\cite{he2021ffb6d}+ICP} 
& 93.1 & 97.0
\\
 \multicolumn{1}{c}{\multirow{-5}{*}{Yes}}     & RCVPose+ICP                         
 & \textbf{95.9} & \textbf{97.2}
 \\ \hline
\end{tabular}
\end{minipage}
\end{table}
\begin{figure*}[t]
\begin{center}
  \includegraphics[width=0.9\textwidth]{
  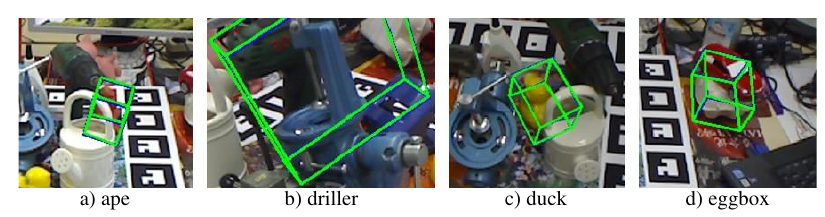}
  \caption{
  RCVPose sample Occlusion LINEMOD results: Blue box = ground truth, green box = estimate. RCVPose shows robustness to (even severe) occlusion\label{fig:some-occluded-LINEMOD-results}}
\end{center}
\end{figure*}

\subsection{Comparison with SOTA}
\label{sec:Results}
We next 
compared
RCVPose against other 
recent competitive methods in the literature.
We achieved state-of-art results on all three datasets, under a moderate training effort (i.e. hyper-parameter adjustment). 
The most challenging dataset was Occlusion LINEMOD,
with results in
Table~\ref{tab:LINEMOD+OccLINEMOD}.
RCVPose+ICP outperformed all other methods on average,
achieving $71.1\%$ mean accuracy,
exceeding the next closest method 
PVN3D by $7.9\%$.
It achieved the top performance on all
objects except duck,
where PVNet had the best result.
Even without ICP refinement, 
RCVPose 
achieved close to the same results at 
$70.2\%$
mean accuracy.

One strength of RCVPose is scale tolerance. Unlike most other methods whose performance reduced with smaller objects, our method 
was not impacted much. 
Significantly, accuracy improved
over FFB6D from $47.2\%$, $45.7\%$ to $61.3\%$, $51.2\%$ for the ape and cat, respectively. 
Another advantage is that it
accumulates votes independently
for each pixel
and is therefore robust to partial occlusions,
capable of
recognizing objects 
that undergo up to $70\%$ occlusion
(see
Fig.~\ref{fig:some-occluded-LINEMOD-results}).
The LINEMOD dataset is less challenging,
as objects are unoccluded.
As listed in Table~\ref{tab:LINEMOD+OccLINEMOD},
RCVPose+ICP still achieved the highest mean accuracy of $99.7\%$,
slightly exceeding the tie between RCVPose (without ICP) and
PVN3D.
RCVPose+ICP was the only method to achieve $100\%$ accuracy for more than one object.
Again the RGB-D methods outperformed all other data modes,
and the top RGB method that included depth refinement~\cite{park2019pix2pose} 
outperformed the best pure RGB method~\cite{Manhardt_2019_ICCV},
supporting the benefits of the added depth mode.

The YCB-Video results in
Table~\ref{tab:ycbvideo}
list AUC and ADD(s),
with and without depth refinement.
RCVPose is the top performing method,
achieving from 
$95.2\%$ to $95.9\%$ ADD(s)
and
from 
$96.6\%$ to $97.2\%$ AUC accuracy,
outperforming the next best method FFB6D
by $2.8\%$ ADD(s)
and $0.2\%$ AUC.
Notably,
RCVPose
increased
ADD(s) of
the relatively small tuna fish can 
by a full  $6\%$ compared to the second best PVN3D.
We also evaluated RCVPose on the BOP challenge benchmark~\cite{hodavn2020bop}, which is a standardized split of a number of 
datasets. Our results on their LINEMOD, Occlusion LINEMOD, and YCB-Video splits showed that RCVPose outperformed all other published results tested on this benchmark, when averaged over all 3 datasets (see Supplementary Material Sec. S.3).
%
%
RCVPose runs at 18 fps on a server with an Intel Xeon 2.3 GHz CPU and RTX8000 GPU for a $640\!\times\!480$ image input. This compares well to other 
voting-based methods,
such as PVNet at 25 fps,
and PVN3D at 5 fps.
The backbone network forward path, radial voting process, and Horn transformation solver take approximately 10, 41, and 4 msecs.  per image respectively at inference time.

\section{Conclusion}
We have proposed RCVPose, a hybrid 6 DoF pose estimator with a 
ResNet-based radial estimator
and a novel keypoint radial voting scheme. 
Our radial voting scheme is shown to be more accurate
than previous schemes,
especially when the keypoints are more dispersed,
which leads to more accurate pose estimation
requiring only 3 keypoints.
We achieved state-of-the-art results on three popular benchmark datasets, YCB-Video, LINEMOD and 
the challenging Occlusion LINEMOD,
ranking high on the BOP Benchmark,
with an 18 fps runtime. 
A limitation is that
training and inference are
executed separately for each object and keypoint (also true for other recent competitive approaches)
and that the 3D voting space is memory intensive,
which will be the focus of future work.
\subsubsection*{Acknowledgements:}
Thanks to Bluewrist Inc. and NSERC for their support of this work.
%
\clearpage
%
%
\bibliographystyle{splncs04}
\bibliography{egbib}
\end{document}


\pagestyle{headings}
\mainmatter
\def\ECCVSubNumber{6108}  

\title{Vote from the Center: 6 DoF Pose Estimation in RGB-D Images by Radial Keypoint Voting \\ 
(Supplementary Material)} 

\titlerunning{Vote from the Center: 6 DoF Pose Estimation in RGB-D Images}
%

\author{Yangzheng Wu\orcidlink{0000-0001-8893-0672}  \and Mohsen Zand\orcidlink{0000-0001-8177-6000
} \and
Ali Etemad\orcidlink{0000-0001-7128-0220} \and
Michael Greenspan\orcidlink{0000-0001-6054-8770}}
\authorrunning{Y. Wu et al.}
%
\institute{Dept. of Electrical and Computer Engineering, 
Ingenuity Labs Research Institute \\ Queen's University, Kingston, Ontario, Canada  }
\maketitle

\section{Overview\label{sec:Overview}}
We document here some addition implementation details, results and further hyperparameter experiments. In Sec.~\ref{sec:ResNet Backbone Structure}, the complete details of the Fully Convolutional ResNet backbone are provided, in sufficient detail to recreate the network. In Sec.~\ref{sec:Detailed Per Category Results of YCB-Video and LINEMOD}, the 6 DoF pose estimation accuracy results for each individual object in the three data sets are presented, as well as some extra bounding box image samples. Finally in Sec.~\ref{sec:Ablation Study}, six additional experiments are included investigating 
the impact of the number of keypoints,
the number of skip connections in the backbone network, 
the use of a combined vs. separate networks for each keypoint, the depth of the backbone network,
the accumulator space resolution, as well as the impact of combining the three different voting schemes into various multi-scheme voting configurations. 
\section{ResNet Backbone Structure\label{sec:ResNet Backbone Structure}}
As shown in Table~\ref{tab:ResStrcture}, we modified ResNet into a Fully Convolutional Network. To start with, we replaced the Fully Connected Layer with a convolutional layer for the following up sampling layers. We then applied up-sampling to the feature map with a combination of convolution, bilinear interpolations, and skip concatenations from the residual blocks. We apply more skip layers than did PVNet~\cite{pvnet},
under the assumption that the convolutional feature maps would preserve more local features than the alternative bilinear interpolation, especially for deeper small scale feature maps.
This design choice was supported by the  experiment described in Sec.~\ref{sec:Number of Skip Connections}.

\begin{table}[ht]
\small
        \caption{ADD(S) metrics for 3 LMO objects
    on different ResNet backbones with ICP. 
    \label{tab:resnets} }
    \begin{center}
    \begin{tabular}{c|c|c|c}\hline
      LMO    & ape & driller & eggbox \\\hline\hline
      ResNet18\_32s  &  60.2   &  77.9   &   81.9     \\\hline
      ResNet34\_32s  & 60.2  &  77.9  &    81.9    \\\hline
      ResNet50\_32s  &  60.8   & 78.4 &    81.9    \\\hline
      ResNet101\_32s &  61.3   & 78.4 &     81.9   \\\hline
      ResNet152\_32s &61.3 & 78.8    &  82.3  \\\hline
    \end{tabular}

    \end{center}
    
\end{table}

We conducted an experiment on three objects, ape, driller and eggbox in Occlusion LINEMOD with different fully convolutional ResNets structures.
Each network is trained until fully convergence with consistent hyper parameter sets.
The resutls are shown in Table~\ref{tab:resnets}.
Deeper ResNet has a tiny performance improvement on three objects tested with a minor sacrifice of speed. 
We ended up with $ResNet 152\_32s$ when conducting the full test on all three datasets.

\section{Accuracy Results Per Object and BOP Benchmark\label{sec:Detailed Per Category Results of YCB-Video and LINEMOD}}
\begin{figure*}[ht]

\begin{center}
  \includegraphics[width=\textwidth]{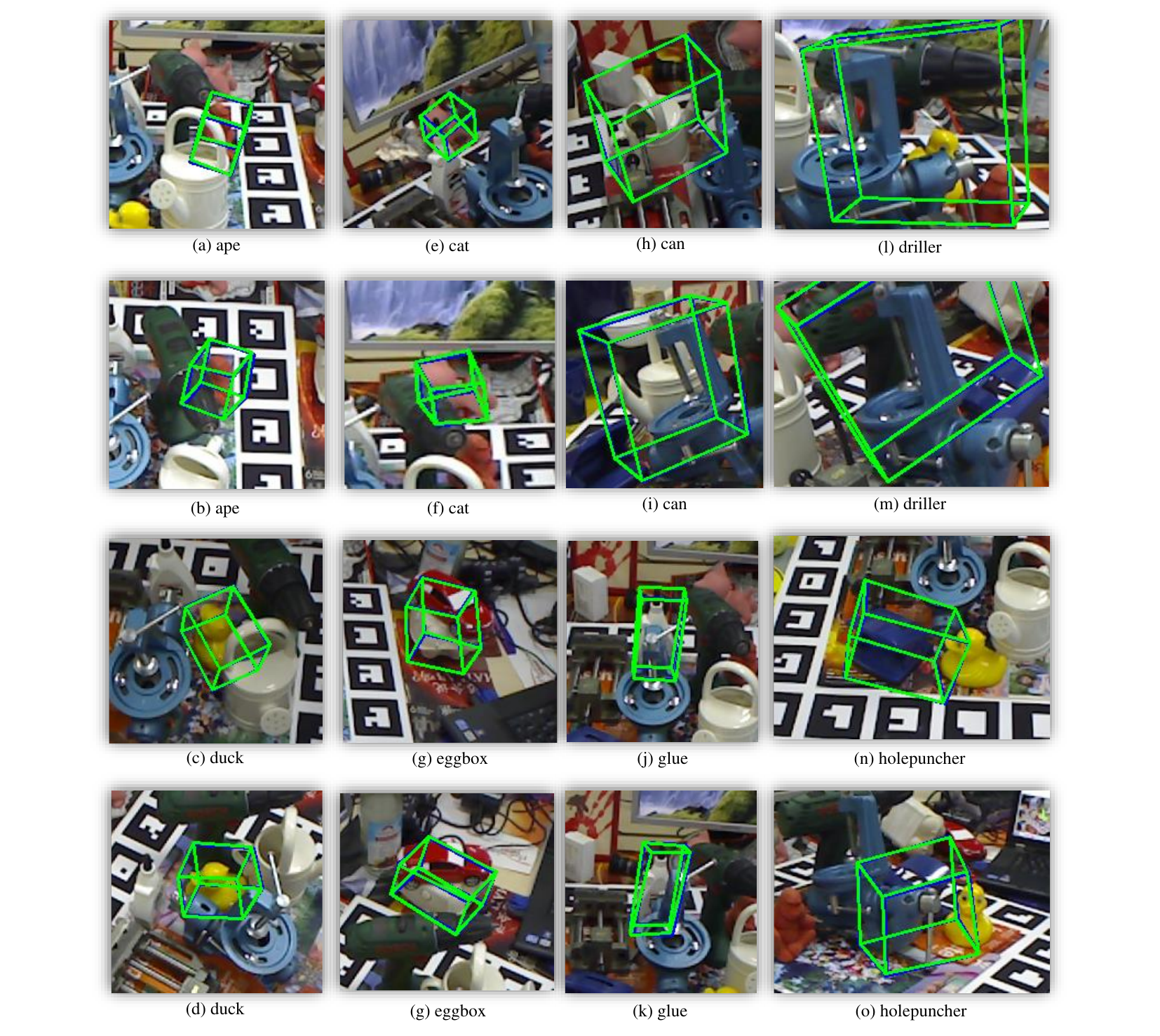}
  \caption{\small Occluded LINEMOD sample results: Blue box = ground truth, green box = estimate\label{fig:all-occluded-LINEMOD-results}}
\end{center}
\end{figure*}

The detailed LINEMOD and
Occlusion LINEMOD ADD(s) results,
and the
YCB-Video ADD(s) and AUC results
categorized per object
are listed in 
Table~\ref{tab:LinemodFull},
Table~\ref{tab:OccLinemod}
and Table~\ref{tab:YCBVideoFull}, respectively.
Some additional successful images
showing recovered and ground truth bounding
boxes are displayed in Figure~\ref{fig:all-occluded-LINEMOD-results}.

As can be seen in
Table~\ref{tab:LinemodFull},
the original LINEMOD
dataset is mostly saturated,
with results from a number of different methods
that are close to perfect. 
Nevertheless,
RCVPose+ICP outperformed all alternatives
at $99.7\%$,
with $100\%$ ADD(s) for three objects,
including the only perfect scores for the
driller and holepuncher objects.

The results in 
Table~\ref{tab:OccLinemod}
show
Occlusion LINEMOD to be quite challenging.
This is not only because of the occluded scenes, but is also due to the fact that the meshes are not very precisely modelled, and that some ground truth poses are not accurate for some cases.

The YCB-Video dataset has two evaluation metrics,
as shown in 
Table~\ref{tab:YCBVideoFull}.
In general, AUC is more foregiving than ADD(s) since AUC has a tolerance of up to 10 cm~\cite{posecnn}. For some objects like the master chef can and the power drill, RCVPose performs slightly worse in AUC compared to PVN3D~\cite{PVN3d}, 
while still performing better in ADD(s). 

All three datasets were also evaluated by the standardized metrics proposed by BOP~\cite{hodavn2020bop}. The results in Table.~\ref{tab:bop} show that our average recall outperformed CosyPose~\cite{labbe2020cosypose} on LINEMOD and occlusion LINEMOD. Although we did not perform better on YCB-Video, we did perform better for the average results over all three datasets. Our method also runs at 18 fps which is also more time efficient compared to 0.36 fps for CosyPose.
\begin{table}[t]
\begin{center}
\caption{\small LINEMOD, Occlusion LINEMOD and YCB-Video evaluated based on BOP Average Recall metrics~\cite{hodavn2020bop}\label{tab:bop}
}
\begin{tabular}{lcccccccccc}
\cline{2-11}
\multicolumn{1}{c}{} &
\multicolumn{3}{c}{$AR_{VSD}$} &
\multicolumn{3}{c}{$AR_{MSSD}$} &
\multicolumn{3}{c}{$AR_{MSPD}$} &
\multicolumn{1}{c}{Average} \\
\cline{1-10}
Method    
& LM & LMO & YCB-V
& LM & LMO & YCB-V
& LM & LMO & YCB-V & $AR$
\\\hline 
PVNet~\cite{pvnet}
& - & 0.43 & -
& - & 0.54 & -
& - & 0.75 & - & -\\
EPOS~\cite{hodan2020epos}
& - & 0.39 & 0.63
& - & 0.50 & 0.68
& - & 0.75 & 0.78 & - \\
SO-Pose~\cite{di2021so}
& - & 0.44 & 0.65
& - & 0.58 & 0.73
& - & 0.82 & 0.76 & - \\ 
CosyPose~\cite{labbe2020cosypose} 
& 0.67 & 0.58 & 0.83
& 0.81 & 0.75 & \bf0.90
& \bf0.84 & \bf0.83 & 0.85 & 0.78\\ 
RCVPose+ICP  
& \bf0.74 & \bf0.68 & \bf0.86
& \bf0.83 & \bf0.77 & 0.86 
& 0.83 & 0.79 & \bf0.86 & \bf0.80\\ \hline
\end{tabular}
\end{center}
\end{table}

\section{Extra Hyperparameter Experiments\label{sec:Extra Ablation Studies}}
\label{sec:Ablation Study}
\noindent
\subsection{Number of Keypoints}
Previous works have used 
between
4~\cite{pavlakos20176},
and up to 
8~\cite{PVN3d,pvnet}
or more~\cite{houghnet} 
keypoints per object,
selected from 
bounding box corners~\cite{rad2017bb8,yolo6d,papandreou2018personlab}
or using the FPS algorithm~\cite{houghnet,PVN3d,pvnet}.
It has been suggested that a greater
number of keypoints is preferable
to improve robustness and accuracy~\cite{PVN3d,pvnet},
especially for pure RGB methods
in which at least 3 keypoints need to be visible for any view of an object to satisfy 
the constraints of the P3P algorithm~\cite{quan1999linear,p3p}.

We examined the impact of
the number of keypoints 
on pose estimation accuracy.
Sets of 3, 4 and 8 keypoints were selected
for the ape, driller and eggbox LINEMOD objects,
using the Bounding Box selection method
described in Sec. 4.5 in the main paper.
The results 
indicate that
increasing the number
of RCVPose keypoints does not impact pose estimation accuracy,
which changed at most only $0.4\%$
between these settings for all three objects.
This is likely due to 
the high accuracy of keypoint location estimation 
under  
radial voting,
which removes the added benefit of 
redundant keypoints.
Given that the
time and memory expense
scale linearly with the number of keypoints,
we settled upon the use of 
the minimal 3 keypoints for 
RCVPose for all of our experiments.
\subsection{Number of Skip Connections\label{sec:Number of Skip Connections}}
There were five different  network
architectures proposed in the intial ResNet paper~\cite{2015Resnet}. 
While some 6 DoF pose recovery works use 
variations of ResNet-18~\cite{pvnet,wang2019densefusion,zakharov2019dpod,trabelsi2021ppn}
others use ResNet-50~\cite{wang2020self6d,Park2020NOL}.
Some customize the structure by converting it to an encoder~\cite{trabelsi2021ppn,Park2020NOL,zakharov2019dpod,wang2019densefusion}, adding extra layers and skip connections~\cite{pvnet} while others use the original ResNet unaltered~\cite{PVN3d,park2019pix2pose}. 
\begin{table}[h]
\begin{center}
\caption{\small Average keypoint estimation error mean ($\mu$ [mm]) and standard deviation ($\sigma$ [mm]) for different ResNet-18 backbone skip connections. Increasing the skip connections reduced the error of the estimation\label{tab:skipVariations}}
\begin{tabular}{ccc|cc}
\cline{2-5}
\multirow{3}{*}{}             & \multicolumn{4}{c}{$\#$ of skip  connections} \\    
                              & \multicolumn{2}{c|}{3} & \multicolumn{2}{c}{5}  \\ \cline{2-5} 
                              & \multicolumn{1}{c}{$\mu$}    & $\sigma$    & $\mu$    & \multicolumn{1}{c}{$\sigma$}      \\ \hline \hline
\multicolumn{1}{c|}{ape}     & 2.4      & 1.1         & 1.8      & \multicolumn{1}{c}{0.8}          \\ \hline
\multicolumn{1}{c|}{driller} & 3.6      & 1.2         & 2.7      & \multicolumn{1}{c}{0.8}           \\ \hline
\multicolumn{1}{c|}{eggbox}  & 3.5      & 1.7         & 2.4      & \multicolumn{1}{c}{1.2}            \\ \hline
\end{tabular}

\end{center}

\end{table}

We conducted an experiment which 
examined the impact of
the number of skip connections
on mean keypoint estimation error
$\bar{\epsilon}$.
We
increased the 
number of skip connections
for ResNet-18,
from 3 to 5.
Such skip connections serve to
improve the influence of image features
during upsampling.
The results are displayed in 
Table~\ref{tab:skipVariations},
and show that increasing 
the skip connections from
3 to 5, decreased 
both the mean
and the standard deviation 
of the keypoint 
estimation error by a large margin,
in all cases. We included 5 skip connections in our architecture, for all experiments, as shown in Fig~\ref{fig:pvVsRCV}.

\subsection{Number of Networks\label{sec:NumberofNetworks}}
Some of the 6 DoF pose estimators trained a single distinct network for each individual object~\cite{PVN3d,pvnet,wang2019densefusion} whereas other multi-class methods trained a single network for all classes combined~\cite{kehl2017ssd,wang2021gdr,di2021so}.
We conducted an experiment on the optimal configuration of the number of networks.
As shown in Table.~\ref{tab:NoofNets},
The radii regression is more accurate when a single network is trained separately on each keypoint compared to training simultaneously on all three keypoints per object.
Therefore, we trained separate networks for each keypoint among each objects to achieve the best performance, with a small sacrifice of the time performance.

\begin{table}[t]
\begin{center}
\caption{\small Keypoint localization error, for training all three keypoints' radii simultaneously in one network and separately in three networks:  
$\bar{\epsilon}$
mean ($\mu_{\{sim|sep\}}$)  and standard deviation ($\sigma_{\{sim|sep\}}$)
for radial voting schemes\label{tab:NoofNets}}
\begin{tabular}{r|cc|cc}

\multicolumn{1}{c}{} &
\multicolumn{4}{c}{$\bar{\epsilon}$ [mm]} \\
\cline{2-5}
\multicolumn{1}{l}{} & \multicolumn{2}{c}{simultaneously}    & \multicolumn{2}{c}{separately} \\[-5pt]
 \multicolumn{1}{c}{}
& \multicolumn{1}{c}{$\mu_{sim}$} & \multicolumn{1}{c}{$\sigma_{sim}$}
& \multicolumn{1}{c}{$\mu_{sep}$} & \multicolumn{1}{c}{$\sigma_{sep}$}\\ \hline \hline
ape & 1.7 & 0.9 & {\bf 1.3} & {\bf 0.7} \\ 
driller & 2.6 & 1.4 & {\bf 2.2} & {\bf 1.0} \\
eggbox & 2.5 & 1.3 & {\bf 2.0} & {\bf 0.7}\\ 
 \hline
\end{tabular}
\end{center}

\end{table}

\subsection{ResNet Backbone Depth\label{sec:ResNet Backbone Depth}} 
\begin{figure}[th]
\begin{center}
\includegraphics[width=1\columnwidth]{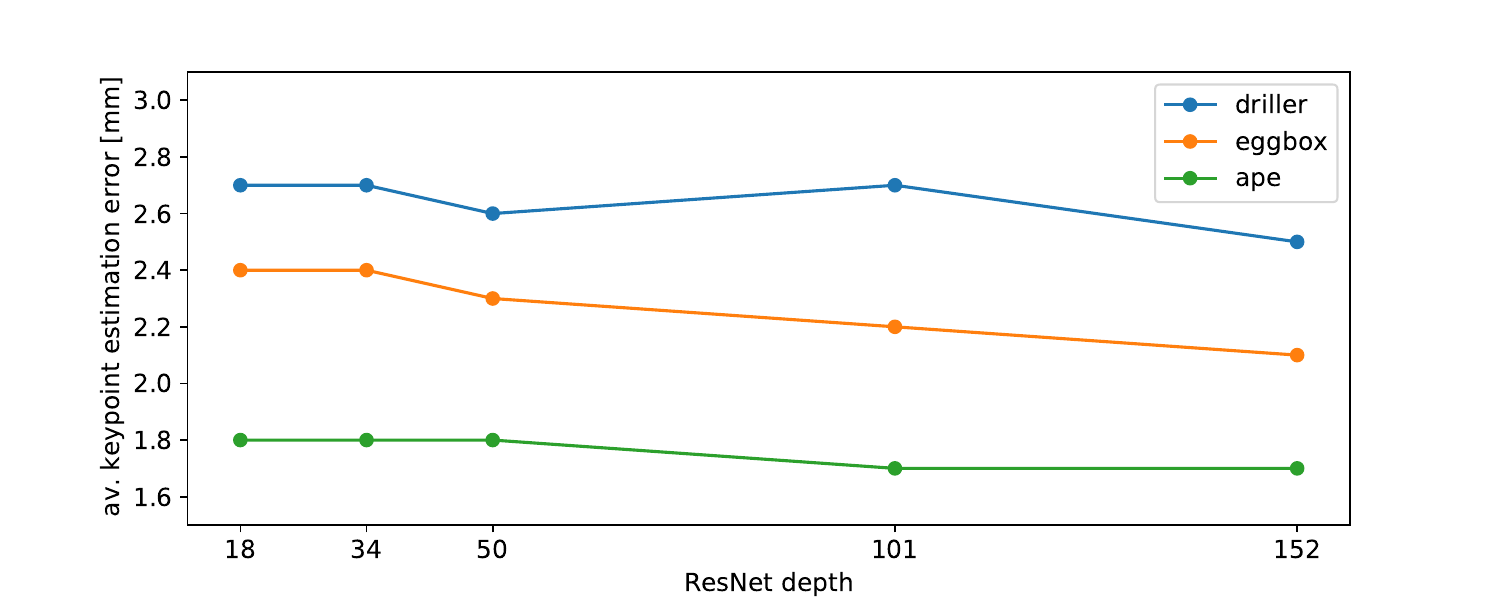}
  \caption{\small Mean keypoint estimation error [mm] vs. ResNet depth
  }
  \label{fig:resnet-depth-plot}
\end{center}
\end{figure}

A further experiment
tested different ResNet depths,
from 18 to 152 layers.
The results are 
plotted in
Fig.~\ref{fig:resnet-depth-plot},
and indicate that the
substantially deeper networks
exhibit only 
a minor reduction in
average keypoint estimation error
$\bar{\epsilon}$.

\begin{figure}[t]
\begin{center}
\includegraphics[width=0.5\columnwidth]{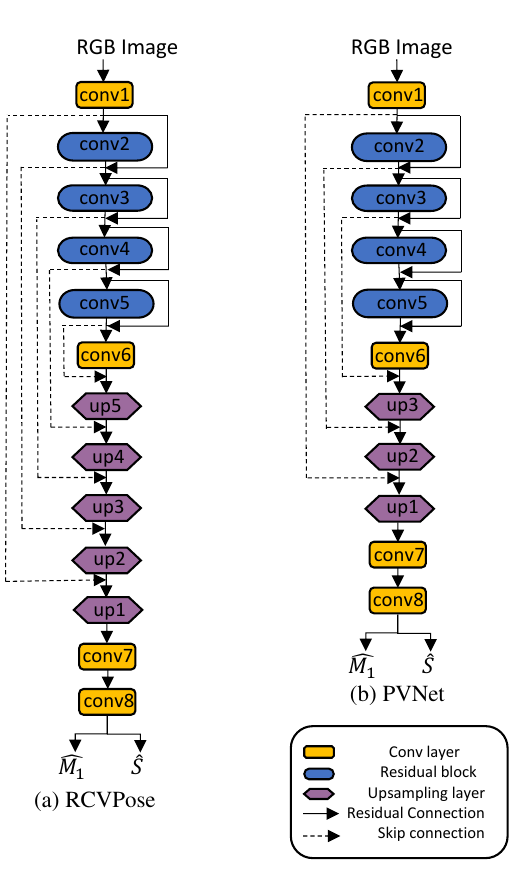}
  \caption{Backbone network structure for (a) RCVPose and (b) PVNet: Denser  skip connections allow more local image features to be kept during upsampling}
  \label{fig:pvVsRCV}
\end{center}
\end{figure}

Despite the rather minor improvement
due to increased depth,
we nevertheless used ResNet-152
with 5 skip connections
in the RCVPose in our experiments, as shown in Fig.~\ref{fig:pvVsRCV} compared to PVNet.
It is likely that we would have received
very similar results had we based
our backbone network on ResNet-18,
albeit with a faster training cycle and
smaller memory footprint.

 \subsection{Accumulator Space Resolution}
 \label{sec:Accumulator Space Resolution}
We varied the
accumulator space resolution
to evaluate the balance of accuracy and efficiency. 
Resolution $\rho$ refers to the linear dimension of a voxel edge 
(i.e. voxel volume $=\rho^3$).
%
%
We selected $6$ different resolutions from $\rho$ = 1 mm to 16 mm,
and ran the voting module for each $\rho$ value with the  same system,
for all 3 scaled bounding box keypoints of all test images of the LINEMOD ape object.

The results are listed in 
Table~\ref{tab:3DAccSpaceResolution}
which shows the
means $\mu_r$ and standard deviations 
$\sigma_r$ of the keypoint estimation errors $\bar{\epsilon}$
and ADD metric, and both the time and space efficiencies,
for varying voxel resolutions.
As expected, 
the voting module was faster and smaller,
and the keypoint estimation error was greater,
at coarser 
resolutions. 
The ADD value,
which is the main metric used to identify a successful
pose estimation event,
remains nearly constant up to a resolution of 5 mm.
The $\rho$ = 5 mm voxel size 
therefore achieved both 
an acceptable speed of 24 fps,
an efficient memory footprint of 3.4 Mbtyes,
and close to the highest ADD value,
and so it was subsequently used throughout the experiments. 


\begin{table}[t]
\begin{center}
\caption{\small Accumulator space resolution $\rho$ [mm] impact on accuracy $\bar{\epsilon}$ [mm], ADD [$\%$], processing speed [fps], and memory [Mbyte], for LINEMOD ape test images. The processing speed includes only the accumulator space time performance
\label{tab:3DAccSpaceResolution}
}
\begin{tabular}{c|c|c|c|c|l}
\hline
$\rho$  & \multicolumn{2}{c|}{$\bar{\epsilon}$ [mm]}  & ADD & speed   &   \multicolumn{1}{|c}{memory}    \\ 
{[mm]} & $\mu_r$   & $\sigma_r$ &[\%]   & [fps]    &    \multicolumn{1}{|c}{[Mbyte]}    \\ 
\hline \hline
0.5              & 1.65                   & 0.63                 & 61.5       & 1.6        & $4\!\times\!957^{3} =3517$\\
1                & 1.75                   & 0.81                 & 61.5       & 5          & $4\!\times\!479^{3} = \ \ 440.61$        \\
2                & 2.33                   & 0.52                  & 61.3       & 12        & $4\!\times\!239^{3} = \ \ \ \ 54.81$      \\
4               & 6.27                    & 0.72                 & 61.3      & 20         & $4\!\times\!118^{3}  = \ \ \ \ \ \ 6.57 $       \\
5               & 6.33                    & 0.69                 & 61.3       & 24          & $4\!\times\!95^{3} \ \ = \ \ \ \ \ \ 3.43 $       \\
8               & 11.73                   & 2.37                 & 55.2       & 32          & $4\!\times\!58^{3} \ \ = \ \ \ \ \ \ 0.78 $       \\
16               & 17.92                   & 5.52                & 45.7        & 40         & $4\!\times\!28^{3} \ \ = \ \ \ \ \ \ 0.09$        \\  \hline
\end{tabular}
\end{center}

\end{table}

\subsection{Ensemble Multi-scheme  Voting\label{sec:Multi-Scheme Voting}} 
The accumulator space is represented exactly the same for
all three voting schemes, 
and is handled in exactly the same manner to
extract keypoint locations through peak detection,
once the voting has been completed.
It is therefore possible and straightforward to combine
voting schemes, by simply adding their resulting accumulator
spaces prior to peak detection. 

We implemented this and compared
the impact of all possible combinations of
offset, vector, and radial voting schemes.
The results are shown in Table~\ref{tab:KEE-BBox-COM},
which also includes the results from each individual voting scheme for comparison. It can be seen that the radial voting scheme outperforms all other alternatives, yielding a lower mean and standard deviation of  keypoint estimation error $\bar{\epsilon}$.
The next best alternative was 
the combination of all three schemes, 
which was greater than 3.5X less accurate than pure radial voting.
Combing radial and offset voting
slightly improved results over pure
offset voting in two of the three objects.
Curiously, combining radial and vector voting degraded results for all objects compared to pure vector voting,
as did combining vector and offset voting.
Based on these results,
it seems possible that there 
may be better ways
than simply adding the 
individual accumulator spaces
to ensemble the information from these
three voting schemes to reduce error further.

\begin{table*}[t]
\begin{center}
\caption{\small Combined Accumulator Space:  
$\bar{\epsilon}$
mean ($\mu_{\{v|o|r\}}$)  and standard deviation ($\sigma_{\{v|o|r\}}$)
for different combination of 3 voting schemes,
with 
$\bar{r}$
= mean distance of keypoints to object centroid
\label{tab:KEE-BBox-COM}
}
\begin{adjustbox}{width=\textwidth}
\begin{tabular}{r|c|cc|cc|cc|cc|cc|cc|cc}
\cline{3-16}
\multicolumn{2}{c}{} &
\multicolumn{10}{c}{$\bar{\epsilon}$ [mm]} \\
\multicolumn{2}{l}{} & \multicolumn{2}{c}{}                & \multicolumn{2}{c}{}                    & \multicolumn{2}{c}{}      & \multicolumn{2}{c}{ vector}  & \multicolumn{2}{c}{}                & \multicolumn{2}{c}{}      & \multicolumn{2}{c}{}             \\ [-3 pt]
\multicolumn{2}{l}{} & \multicolumn{2}{c}{vector}                & \multicolumn{2}{c}{vector}                    & \multicolumn{2}{c}{radial}      & \multicolumn{2}{c}{ + offset}  & \multicolumn{2}{c}{}                & \multicolumn{2}{c}{}      & \multicolumn{2}{c}{}             \\[-3 pt]
\multicolumn{2}{l}{} & \multicolumn{2}{c}{+ offset}                & \multicolumn{2}{c}{+ radial}                    & \multicolumn{2}{c}{+ offset}      & \multicolumn{2}{c}{ + radial}  & \multicolumn{2}{c}{vector}                & \multicolumn{2}{c}{offset}      & \multicolumn{2}{c}{radial}             \\ [-3 pt]
\cline{2-2} 
 \multicolumn{1}{c}{}
 &
 \multicolumn{1}{c|}{$\bar{r}$ [mm]}
&
\multicolumn{1}{c}{$\mu_v$} & \multicolumn{1}{c}{$\sigma_v$} & \multicolumn{1}{|c}{$\mu_o$} & \multicolumn{1}{c|}{$\sigma_o$} & \multicolumn{1}{c}{$\mu_r$} & \multicolumn{1}{c}{$\sigma_r$} & \multicolumn{1}{|c}{$\mu_r$} & \multicolumn{1}{c}{$\sigma_r$} & \multicolumn{1}{|c}{$\mu_r$} & \multicolumn{1}{c}{$\sigma_r$} & \multicolumn{1}{|c}{$\mu_r$} & \multicolumn{1}{c}{$\sigma_r$} & \multicolumn{1}{|c}{$\mu_r$} & \multicolumn{1}{c}{$\sigma_r$}  \\ \hline \hline
 ape & 142.1 & 20.2 & 12.4 & 12.7 & 6.7 & 9.8 & 6.2  & 7.2 & 1.2 & 12.5 & 7.6 & 10.4 & 5.3 & 1.8 & 0.8 \\ \hline
driller & 318.8 & 22.3 & 11.7 & 13.3 & 7.9  & 8.7 & 3.4 & 5.7 & 2.3 & 11.3 &  8.2 &  9.5 &  3.5  & 2.7 & 0.8 \\ \hline
eggbox & 197.3 & 21.6 & 13.5  & 17.4  & 10.5  & 12.1 & 5.2 & 6.4 & 3.3 & 13.7 &  8.5 &  11.4 &  4.7 & 2.4 & 1.2 \\ 
 \hline
\end{tabular}
\end{adjustbox}
\end{center}

\end{table*}

\begin{table*}[t]
\begin{center}
\caption{\small Occlusion LINEMOD Accuracy Results.
Non-symmetric objects are evaluated with ADD, and
symmetric objects
(annotated with~$^{*}$) are evaluated with ADD-s
\label{tab:OccLinemod}
}
\resizebox{1.0\columnwidth}{!}{%
\begin{tabular}{c|c|c|c|c|c|c|c|c|c|c}
\cline{3-10}
\multicolumn{2}{c|}{}
&
\multicolumn{8}{c|}{Object}  \\ [3pt]
\cline{1-2}
\cline{11-11}
Mode      & Method     & ape  & can  & cat  & driller & duck & eggbox$^{*}$ & glue$^{*}$ & holepuncher &
\begin{tabular}{c}
$\overline{\mbox{ADD(s)}} [\%]$ 
\end{tabular}
\\\hline \hline
                                & Oberweger~\cite{oberweger2018making}  & 12.1 & 39.9 & 8.2  & 45.2    & 17.2 & 22.1   & 35.8 & 36.0        & 27.1 \\\cline{2-11} 
                                & Hu et al.~\cite{hu2019segmentation}  & 17.6 & 53.9 & 3.3  & 62.4    & 19.2 & 25.9   & 39.6 & 21.3        & 30.4 \\\cline{2-11} 
                                & Pix2Pose~\cite{park2019pix2pose}   & 22.0 & 44.7 & 22.7 & 44.7    & 15.0 & 25.2   & 32.4 & 49.5        & 32.0 \\\cline{2-11} 
                                & DPOD~\cite{zakharov2019dpod}       & -  & -  & -  & -     & -  & -    & -  & -         & 32.8 \\\cline{2-11} 
                                & PVNet~\cite{pvnet}      & 15.8 & 63.3 & 16.7 & 25.2    & \textbf{65.7} & 50.2   & 49.6 & 39.7        & 40.8 \\\cline{2-11} 
\multirow{-6}{*}{RGB}
& PPRN~\cite{trabelsi2021ppn}       & -  & -  & -  & -     & -  & -    & -  & -         & 58.4 \\\hline\hline
                                & YOLO6D~\cite{yolo6d}     & -  & -  & -  & -     & -  & -    & -  & -         & 6.4  \\\cline{2-11} 
                                & SSD6D+ref~\cite{kehl2017ssd}  & -  & -  & -  & -     & -  & -    & -  & -         & 27.5 \\\cline{2-11} 
                               
\multirow{-3}{*}{RGB}      
&  PoseCNN~\cite{posecnn}    & 9.6  & 45.2 & 0.9  & 41.4    & 19.6 & 22.0   & 38.5 & 22.1        & 24.9 \\\cline{2-11} 
\multirow{-3}{*}{+D ref}
& DPOD+ref~\cite{zakharov2019dpod}   & -  & -  & -  & -     & -  & -    & -  & -         & 47.3 \\\hline\hline
                                & PVN3D~\cite{PVN3d}      & 33.9 & 88.6 & 39.1 & 78.4   & 41.9 & 80.9   & 68.1 & 74.7        & 63.2 \\\cline{2-11} 
                                & RCVPose     & 60.3 & 92.5 & 50.2 & 78.2    & 52.1 & 81.2   & 72.1  & 75.2        & 70.2 \\\cline{2-11} 
\multirow{-3}{*}{RGB-D}  
& RCVPose+ICP & {\bf 61.3} & {\bf 93} & {\bf 51.2} & {\bf 78.8}   & 53.4 & {\bf 82.3}   & {\bf 72.9} & {\bf 75.8}        & {\bf 71.1}\\ \hline
\end{tabular}
} 
\end{center}

\end{table*}

\begin{sidewaystable}[]
%
\begin{center}
\caption{\small LINEMOD Accuracy Results: Non-symmetric objects are evaluated with ADD, and
symmetric objects
(annotated with~$^{*}$) are evaluated with ADD-s
\label{tab:LinemodFull}
}
\resizebox{1.0\columnwidth}{!}{%
\begin{tabular}{|c|c|c|c|c|c|c|c|c|c|c|c|c|c|c|c|c|}
\cline{3-16}
\multicolumn{2}{c|}{}
&
\multicolumn{14}{c|}{
Object}  \\
\multicolumn{2}{c|}{}
&
\multicolumn{1}{c}{}  &
\multicolumn{1}{c}{bench-}  &
\multicolumn{7}{c}{}  &
\multicolumn{1}{c}{hole-}  &
\multicolumn{4}{c|}{} \\
\cline{1-2}
Mode      & Method     & ape & vise & camera  & can  & cat  & driller & duck & eggbox$^{*}$ & glue$^{*}$ & puncher & iron & lamp & phone & mean \\\hline \hline
                                & SSD6D+ref~\cite{kehl2017ssd}  & 2.6  & 15.1 & 6.1  & 27.3  & 9.3     & 12.0  & 1.3 & 2.8  & 3.4 & 3.1  & 14.6    & 11.4  & 9.7        & 9.1 \\\cline{2-16}  
                                & Pix2Pose~\cite{park2019pix2pose}   & 58.1 & 91 & 60.9 & 84.4    & 65 & 76.3   & 43.8 & 96.8 & 79.4 & 74.8 & 83.4 & 82 & 45        & 72.4 \\\cline{2-16}  
                                & DPOD~\cite{zakharov2019dpod}       & 53.3  & 95.3  & 90.4  & 94.1     & 60.4  & 97.7    & 66  & 99.7 & 93.8 & 65.8 & 99.8 & 88.1 & 74.2          & 83.0 \\\cline{2-16}  
                                & PVNet~\cite{pvnet}      & 43.62 & 99.9 & 86.9 & 95.5    & 79.3 & 96.4   & 52.6 & 99.2        & 95.7 & 81.9 & 98.9 & 99.3 & 92.4 & 86.3 \\\cline{2-16}  
                                & PPRN~\cite{trabelsi2021ppn}       & 84.5  & 98.7  & 93.7  & 97.8     & 87.3  & 96.9    & 88.5  & 98.5 & 99.5 & 84.5 & 99.1 & 98.7 & 92.5         & 93.9 \\\cline{2-16}  
\multirow{-6}{*}{RGB}           & DeepIM~\cite{li2018deepim}                  & 77       & 97.5       & 93.5        & 96.5           & 82.1         & 95.0        & 77.7       & 97.1       & 99.4          & 52.8       & 98.3        & 97.5        & 87.7       & 88.6 \\\hline\hline
                                & YOLO6D~\cite{yolo6d}     & 21.6  & 81.8  & 36.6  & 68.8     & 41.8  & 63.5    & 27.2  & 69.6 & 80 & 42.6 & 75 & 71.1 & 47.7 & 56.0  \\\cline{2-16} 
\multirow{-2}{*}{RGB}
                                & SSD6D+ref~\cite{kehl2017ssd}  & -  & - & -  & -  & -     & -  & -    & -  & - & -  & -    & -  & -        & 34.1 \\\cline{2-16}  
\multirow{-2}{*}{+D ref}      & DPOD+ref~\cite{zakharov2019dpod}  & 87.7       & 98.5        & 96.1        & 99.7   & 94.7   & 98.8        & 86.3       & 99.9       &  96.8         & 86.8      & 100        & 96.8  &  94.7      & 95.2 \\\hline \hline
                                & DenseFusion~\cite{wang2019densefusion}      & 92.3       & 93.2        & 94.4        & 93.1   & 96.5   & 87.0        & 92.3       & 99.8       &  100.0         & 92.1      & 97.0        & 95.3  &  92.8      & 94.3 \\\cline{2-16}  
                                & PVN3D~\cite{PVN3d}      & 97.3       & 99.7        & 99.6        & {\bf 99.5}   & {\bf 99.8}   & 99.3        & 98.2       & 99.8       &  100.0         & 99.9       & 99.7        & {\bf 99.8} &  99.5      & 99.4 \\\cline{2-16}  
                                & RCVPose                  & 99.2       & 99.6        & 99.7        & 99           & 99.4         & 99.7        & 99.4       & 98.7       & 99.7           & 99.8       & 99.9        & 99.2        & 99.1       & 99.43 \\\cline{2-16}  
\multirow{-4}{*}{RGB-D}         & RCVPose+ICP              & {\bf 99.6} & {\bf 99.7}  & {\bf 99.7}  & 99.3         & 99.7         & {\bf 100}   & {\bf 99.7} & 99.3       & {\bf 100.0}    & {\bf 100}  & {\bf 99.9}  & 99.5       & {\bf99.7}  & {\bf 99.7}\\ \hline
\end{tabular}
} 
\end{center}

\vspace{0.5\baselineskip}

\end{sidewaystable}

\begin{sidewaystable}[]
\begin{center}
\caption{\small YCB Video AUC~\cite{posecnn} and ADD(s)~\cite{hinterstoisser2012model} results:Non-symmetric objects are evaluated with ADD, and
symmetric objects
(annotated with~$^{*}$) are evaluated with ADD-s. The AUC metrics is based on the curve with ADD for non-symmetries and ADDs with symmetries
\label{tab:YCBVideoFull}
}
\resizebox{1.0\columnwidth}{!}{%
\begin{tabular}{|ccc|c|c|c|c|c|c|c|c|c|c|c|c|c|c|c|c|c|c|c|c|c|c|}
   \multicolumn{25}{c}{}  \\
\multicolumn{1}{c}{}&&&
\begin{rotate}{60}
\hspace{-0.65 cm} 002 master
\end{rotate}
&
\begin{rotate}{60}
\hspace{-0.65 cm} 003 cracker
\end{rotate}
&
\begin{rotate}{60}
\hspace{-0.65 cm} 004 sugar
\end{rotate}
&
\begin{rotate}{60}
\hspace{-0.65 cm} 005 tomato  
\end{rotate}
&
\begin{rotate}{60}
\hspace{-0.65 cm} 006 mustard  
\end{rotate}
&
\begin{rotate}{60}
\hspace{-0.65 cm} 007 tuna fish  
\end{rotate}
&
\begin{rotate}{60}
\hspace{-0.65 cm} 008 pudding 
\end{rotate}
&
\begin{rotate}{60}
\hspace{-0.65 cm} 009 gelatin
\end{rotate}
&
\begin{rotate}{60}
\hspace{-0.65 cm} 010 potted  
\end{rotate}
&
\begin{rotate}{60}
\hspace{-0.65 cm} 011 banana  
\end{rotate}
&
\begin{rotate}{60}
\hspace{-0.65 cm} 019 pitcher 
\end{rotate}
&
\begin{rotate}{60}
\hspace{-0.65 cm} 021 bleach 
\end{rotate}
&
\begin{rotate}{60}
\hspace{-0.65 cm} 024 bowl$^{*}$ 
\end{rotate}
&
\begin{rotate}{60}
\hspace{-0.65 cm} 025 mug 
\end{rotate}
&
\begin{rotate}{60}
\hspace{-0.65 cm} 035 power 
\end{rotate}
&
\begin{rotate}{60}
\hspace{-0.65 cm} 036 wood 
\end{rotate}
&
\begin{rotate}{ 60}
\hspace{-0.65 cm} 037 scissors 
\end{rotate}
&
\begin{rotate}{60}
\hspace{-0.65 cm} 040 large 
\end{rotate}
&
\begin{rotate}{60}
\hspace{-0.65 cm} 051 large 
\end{rotate}
&
\begin{rotate}{60}
\hspace{-0.65 cm} 052 extra large$^{*}$ 
\end{rotate}
&
\begin{rotate}{60}
\hspace{-0.65 cm} 061 foam 
\end{rotate}
&\\

\cline{1-3} 
\multicolumn{1}{|c}{Refine} & \multicolumn{1}{|c}{Metric}  & \multicolumn{1}{|c|}{Method}                         & 
\begin{rotate}{60}
\hspace{0.8 cm} {chef can}
\end{rotate}

& 
\begin{rotate}{60}
\hspace{0.8 cm}
{box} 
\end{rotate}
&
\begin{rotate}{60}
\hspace{0.8 cm}{box} 
\end{rotate}
&
\begin{rotate}{60}
\hspace{0.8 cm}{soup can} 
\end{rotate}
&
\begin{rotate}{60}
\hspace{0.8 cm}{bottle} 
\end{rotate}
&
\begin{rotate}{60}
\hspace{0.8 cm}{can} 
\end{rotate}
&
\begin{rotate}{60}
\hspace{0.8 cm}{box} 
\end{rotate}
&
\begin{rotate}{60}
\hspace{0.8 cm}{box} 
\end{rotate}
&
\begin{rotate}{60}
\hspace{0.8 cm}{meat can} 
\end{rotate}
&
&
\begin{rotate}{60}
\hspace{0.8 cm}{base} 
\end{rotate}
&
\begin{rotate}{60}
\hspace{0.8 cm}{cleanser} 
\end{rotate}
&
&
&
\begin{rotate}{60}
\hspace{0.8 cm}{drill} 
\end{rotate}
&
\begin{rotate}{60}
\hspace{0.8 cm}{block$^{*}$} 
\end{rotate}
&
&
\begin{rotate}{60}
\hspace{0.8 cm}{marker} 
\end{rotate}
&
\begin{rotate}{60}
\hspace{0.8 cm}{clamp$^{*}$} 
\end{rotate}
&
\begin{rotate}{60}
\hspace{0.8 cm}{clamp}
\end{rotate}
&
\begin{rotate}{60}
\hspace{0.8 cm}{brick$^{*}$} 
\end{rotate}
&
{mean}   
\\ \hline \hline
\multicolumn{1}{|c|}{{}}       & \multicolumn{1}{c|}{}                          & {PoseCNN~\cite{posecnn} }       & {83.9}                & {76.9}            & {84.2}          & {81.0}                & {90.4}               & {88.0}              & {79.1}            & {87.2}            & {78.5}                & {86.0}       & {77.0}             & {71.6}                & {69.6}     & {78.2}    & {72.7}            & {64.3}           & {56.9}          & {71.7}             & {50.2}            & { 44.1}                  & {88.0}           & {75.8}          \\ \cline{3-25} 
\multicolumn{1}{|c|}{{}}       & \multicolumn{1}{c|}{}                          & {DF(per-pixel)\cite{wang2019densefusion}} & {95.3}                & {92.5}            & {95.1}          & {93.8}                & {95.8}               & {95.7}              & {94.3}            & {97.2}            & {89.3}                & {90.0}       & {93.6}             & {94.4}                & {86.0}     & {95.3}    & {92.1}            & {89.5}           & {90.1}          & {95.1}             & {71.5}            & {70.2}                  & {92.2}           & {91.2}          \\ \cline{3-25} 
\multicolumn{1}{|c|}{{}}       & \multicolumn{1}{c|}{}                          & {PVN3D\cite{PVN3d}}         & { \textbf{96.0}}       & { 96.1}            & { 97.4}          & { 96.2}                & { 97.5}               & 96                                       & { 97.1}            & { 97.7}            & { 93.3}                & 96.6                              & { \textbf{97.4}}    & { 96.0}                & {90.2}     & { 97.6}    & { \textbf{96.7}}   & {\textbf{90.4}}  & { \textbf{96.7}} & { \textbf{96.7}}    & { 93.6}            & {88.4}                  & { \textbf{96.8}}  & { \textit{95.5}} \\ \cline{3-25} 
\multicolumn{1}{|c|}{{}}       & \multicolumn{1}{c|}{\multirow{-4}{*}{AUC}}     & RCVPose                               & 95.7                                       & \textbf{97.2}                          & \textbf{97.6}                        & \textbf{98.2}                              & \textbf{97.9}                             & \textbf{98.2}                            & \textbf{97.7}                          & 97.7                                   & \textbf{97.9}                              & \textbf{97.9}                     & 96.2                                    & \textbf{99.2}                              & \textbf{95.2}                   & \textbf{98.4}                  & 96.2                                   & 89.1                                  & 96.2                                 & 95.9                                    & \textbf{95.2}   & \textbf{94.7}      & 95.7                                  & \textbf{96.6}                    \\ \cline{2-25} 
\multicolumn{1}{|c|}{{}}       & \multicolumn{1}{c|}{}                          & {PoseCNN~\cite{posecnn} }       & {50.2}                & {53.1}            & {68.4}          & {66.2}                & {81.0}               & {70.7}              & {62.7}            & {75.2}            & {\textit{59.5}}       & {72.3}       & {53.3}             & {50.3}                & {69.6}     & {58.5}    & {55.3}            & {64.3}           & {35.8}          & {58.3}             & {50.2}            & { 44.1}                  & {88.0}           & {59.9}          \\ \cline{3-25}
\multicolumn{1}{|c|}{\multirow{-4}{*}{No}}       
& \multicolumn{1}{c|}{}                          & {DF(per-pixel)\cite{wang2019densefusion}} & {70.7}                & {86.9}            & {90.8}          & {84.7}                & {90.9}               & {79.6}              & {89.3}            & {95.8}            & {79.6}                & {76.7}       & {87.1}             & {87.5}                & {86.0}     & {83.8}    & {83.7}            & {89.5}           & {77.4}          & {89.1}             & {71.5}            & {70.2}                  & {92.2}           & {82.9}          \\ \cline{3-25} 
& \multicolumn{1}{|c|}{}                          & {PVN3D\cite{PVN3d}}         & { 80.5}                & { 94.8}            & 96.3                                 & {88.5}                & 96.2                                      & {89.3}              & { 95.7}            & 96.1                                   & { 88.6}                & { 93.7}       & \textbf{96.5}                           & { 93.2}                & {90.2}     & { 95.4}    & { 95.1}            & {\textbf{90.4}}  & { 92.7}          & { 91.8}             & { 93.6}            & {88.4}                  & { \textbf{96.8}}  & { 91.8}          \\ \cline{3-25} 
& \multicolumn{1}{|c|}{\multirow{-4}{*}{ADD (s)}} & RCVPose                               & \textbf{93.6}                              & \textbf{95.7}                          & \textbf{97.2}                        & \textbf{94.7}                              & \textbf{97.2}                             & \textbf{96.4}                            & \textbf{97.1}                          & \textbf{96.5}                          & \textbf{90.2}                              & \textbf{96.7}                     & 95.7                                    & \textbf{97.8}                              & \textbf{94.9}                   & \textbf{96.3}                  & \textbf{95.4}                          & 89.3                                  & \textbf{94.7}                        & \textbf{92.4}                           & \textbf{96.4} & \textbf{94.7}      & 95.7                                  & \textbf{95.2}                    \\ \hline \hline
\multicolumn{1}{|c|}{{}}       & \multicolumn{1}{c|}{}                          & {PoseCNN~\cite{posecnn} +ICP}   & { 95.8}                & { 92.7}            & { 98.2}          & { 94.5}                & { 98.6}               & { 97.1}              & { 97.9}            & { 98.8}            & { 92.7}                & { 97.1}       & { 97.8}             & { 96.9}                & {81.0}     & { 94.9}    & { 98.2}            & {87.6}           & { 91.7}          & { 97.2}             & {75.2}            & {64.4}                  & { 97.2}           & { 93.0}          \\ \cline{3-25} 
\multicolumn{1}{|c|}{{}}       & \multicolumn{1}{c|}{}                          & {DF(iterative)\cite{wang2019densefusion}} & { 96.4}                & { 95.8}            & { 97.6}          & { 94.5}                & { 97.3}               & { 97.1}              & { 96.0}            & { 98.0}            & { 90.7}                & { 96.2}       & { 97.5}             & { 95.9}                & {89.5}     & { 96.7}    & { 96.0}            & { 92.8}           & { 92.0}          & { 97.6}             & {72.5}            & { 69.9}                  & {92.0}           & { 93.2}          \\ \cline{3-25} 
\multicolumn{1}{|c|}{{}}       & \multicolumn{1}{c|}{}                          & {PVN3D\cite{PVN3d}+ICP}     & { 95.2}                & {94.4}            & {97.9}          & { 95.9}                & {\textbf{98.3}}      & {96.7}              & { \textbf{98.2}}   & { \textbf{98.8}}   & { 93.8}                & { 98.2}       & {\textbf{97.6}}    & { 97.2}                & { 92.8}     & { 97.7}    & {\textbf{97. 1}}  & {\textbf{91.1}}  & {95.0}          & { 98.1}             & { 95.6}            & { 90.5}                  & { \textbf{98.2}}  & { 96.1}          \\ \cline{3-25} 
\multicolumn{1}{|c|}{{}}       & \multicolumn{1}{c|}{\multirow{-4}{*}{AUC}}     & RCVPose+ICP                           & \textbf{96.2}                              & \textbf{97.9}                          & 97.9                                 & \textbf{99}                                & 98.2                                      & \textbf{98.6}                            & 98.1                                   & 98.4                                   & \textbf{98.4}                              & \textbf{98.3}                     & 97.2                                    & \textbf{99.6}                              & \textbf{96.9}                   & \textbf{98.7}                  & 96.4                                   & 90.7                                  & \textbf{96.4}                        & \textbf{96.6}                           & \textbf{96.2}   & \textbf{95.1}      & 96.6                                  & \textbf{97.2}                    \\ \cline{2-25} 
\multicolumn{1}{|c|}{{}}       & \multicolumn{1}{c|}{}                          & {PoseCNN~\cite{posecnn} +ICP}   & { 68.1}                & {83.4}            & { 97.1}          & {81.8}                & { 98.0}               & {83.9}              & { 96.6}            & { 98.1}            & {83.5}                & { 91.9}       & { 96.9}             & { 92.5}                & {81.0}     & {81.1}    & { 97.7}            & {87.6}           & {78.4}          & {85.3}             & {75.2}            & { 64.4}                  & { 97.2}           & {85.4}          \\ \cline{3-25} 
\multicolumn{1}{|c|}{\multirow{-4}{*}{Yes}}       & \multicolumn{1}{c|}{}                          & {DF(iterative)\cite{wang2019densefusion}} & {73.2}                & { 94.1}            & { 96.5}          & {85.5}                & { 94.7}               & {81.9}              & { 93.3}            & { 96.7}            & {83.6}                & {83.3}       & { 96.9}             & {89.9}                & {89.5}     & {88.9}    & { 92.7}            & { 92.8}           & {77.9}          & { 93.0}             & {72.5}            & {69.9}                  & { 92.0}           & {86.1}          \\ \cline{3-25} 
& \multicolumn{1}{|c|}{}                          & {PVN3D\cite{PVN3d}+ICP}     & {79.3}                & { 91.5}            & { 96.9}          & { 89.0}                & { \textbf{97.9}}      & { 90.7}              & { 97.1}            & { \textbf{98.3}}   & {87.9}                & { 96.0}       & { \textbf{96.9}}    & { 95.9}                & { 92.8}     & { 96.0}    & { 95.7}            & { 91.1}           & {87.2}          & { 91.6}             & { 95.6}            & { 90.5}                  & { \textbf{98.2}}  & { 92.3}          \\ \cline{3-25} 
& \multicolumn{1}{|c|}{\multirow{-4}{*}{ADD (s)}} & RCVPose+ICP                           & \textbf{94.7}                              & \textbf{96.4}                          & \textbf{97.6}                        & \textbf{95.4}                              & 97.7                                      & \textbf{96.7}                            & \textbf{97.4}                          & 97.9                                   & \textbf{92.6}                              & \textbf{97.2}                     & 96.7                                    & \textbf{98.4}                              & \textbf{95.3}                   & \textbf{97.1}                  & \textbf{96.2}                          & \textbf{91.2}                         & \textbf{94.9}                        & \textbf{93.2}                           & \textbf{96.7}    & \textbf{94.9}      & 96.6                                  & \textbf{95.9}                    \\ \hline
\end{tabular}
}
\end{center}

\end{sidewaystable}

\begin{table*}[ht]
\begin{center}
\caption{\small ResNet Backbone structure compared to PVNet
\label{tab:ResStrcture}
}
\begin{adjustbox}{max width=\textwidth}
\begin{tabular}{ccccccc}
\cline{1-7}
\multicolumn{7}{c}{ResNet Backbone Strcture}\\ \cline{1-7}
\multicolumn{1}{c|}{Layer}                  & \multicolumn{1}{c|}{ResNet-152 32s(RCVPose)}                                                                                 & \multicolumn{1}{c|}{ResNet-101 32s}                                                                                          & \multicolumn{1}{c|}{ResNet-50 32s}                                                                                          & \multicolumn{1}{c|}{ResNet-34 32s}                                                                       & \multicolumn{1}{c|}{ResNet-18 32s}                                                                       & \multicolumn{1}{c}{ResNet-18 8s(PVNet)}                                                                            \\ \hline\hline
\multicolumn{1}{c|}{conv1}                  & \multicolumn{6}{c}{$7 \times 7$,  $64$,  stride $2$}  \\ \cline{1-7}
\multicolumn{1}{c|}{\multirow{2}{*}{conv2}} & \multicolumn{6}{c}{$3 \times 3$ max pool, stride $2$} \\ \cline{2-7}
\multicolumn{1}{c|}{}                   & \multicolumn{1}{c|}{$\left[\begin{array}{c}1 \times 1,64 \\ 3 \times 3,64 \\ 1 \times 1,256\end{array}\right] \times 3$}     & \multicolumn{1}{c|}{$\left[\begin{array}{c}1 \times 1,64 \\ 3 \times 3,64 \\ 1 \times 1,256\end{array}\right] \times 3$}     & \multicolumn{1}{c|}{$\left[\begin{array}{c}1 \times 1,64 \\ 3 \times 3,64 \\ 1 \times 1,256\end{array}\right] \times 3$}    & \multicolumn{1}{c|}{$\left[\begin{array}{l}3 \times 3,64 \\ 3 \times 3,64\end{array}\right] \times 3$}   & \multicolumn{1}{c|}{$\left[\begin{array}{l}3 \times 3,64 \\ 3 \times 3,64\end{array}\right] \times 2$}   & \multicolumn{1}{c}{$\left[\begin{array}{l}3 \times 3,64 \\ 3 \times 3,64\end{array}\right] \times 2$}  \\ \cline{1-7}
\multicolumn{1}{c|}{conv3}                  & \multicolumn{1}{c|}{$\left[\begin{array}{c}1 \times 1,128 \\ 3 \times 3,128 \\ 1 \times 1,512\end{array}\right] \times 8$}   & \multicolumn{1}{c|}{$\left[\begin{array}{c}1 \times 1,128 \\ 3 \times 3,128 \\ 1 \times 1,512\end{array}\right] \times 4$}   & \multicolumn{1}{c|}{$\left[\begin{array}{c}1 \times 1,128 \\ 3 \times 3,128 \\ 1 \times 1,512\end{array}\right] \times 4$}  & \multicolumn{1}{c|}{$\left[\begin{array}{l}3 \times 3,128 \\ 3 \times 3,128\end{array}\right] \times 4$} & \multicolumn{1}{c|}{$\left[\begin{array}{l}3 \times 3,128 \\ 3 \times 3,128\end{array}\right] \times 2$} & \multicolumn{1}{c}{$\left[\begin{array}{l}3 \times 3,128 \\ 3 \times 3,128\end{array}\right] \times 2$}    \\ \cline{1-7}
\multicolumn{1}{c|}{conv4}                  & \multicolumn{1}{c|}{$\left[\begin{array}{c}1 \times 1,256 \\ 3 \times 3,256 \\ 1 \times 1,1024\end{array}\right] \times 36$} & \multicolumn{1}{c|}{$\left[\begin{array}{c}1 \times 1,256 \\ 3 \times 3,256 \\ 1 \times 1,1024\end{array}\right] \times 23$} & \multicolumn{1}{c|}{$\left[\begin{array}{c}1 \times 1,256 \\ 3 \times 3,256 \\ 1 \times 1,1024\end{array}\right] \times 6$} & \multicolumn{1}{c|}{$\left[\begin{array}{l}3 \times 3,256 \\ 3 \times 3,256\end{array}\right] \times 6$} & \multicolumn{1}{c|}{$\left[\begin{array}{l}3 \times 3,256 \\ 3 \times 3,256\end{array}\right] \times 2$} & \multicolumn{1}{c}{$\left[\begin{array}{l}3 \times 3,256 \\ 3 \times 3,256\end{array}\right] \times 2$} \\ \cline{1-7}
\multicolumn{1}{c|}{conv5}                  & \multicolumn{1}{c|}{$\left[\begin{array}{c}1 \times 1,512 \\ 3 \times 3,512 \\ 1 \times 1,2048\end{array}\right] \times 3$}  & \multicolumn{1}{c|}{$\left[\begin{array}{c}1 \times 1,512 \\ 3 \times 3,512 \\ 1 \times 1,2048\end{array}\right] \times 3$}  & \multicolumn{1}{c|}{$\left[\begin{array}{c}1 \times 1,512 \\ 3 \times 3,512 \\ 1 \times 1,2048\end{array}\right] \times 3$} & \multicolumn{1}{c|}{$\left[\begin{array}{l}3 \times 3,512 \\ 3 \times 3,512\end{array}\right] \times 3$} & \multicolumn{1}{c|}{$\left[\begin{array}{l}3 \times 3,512 \\ 3 \times 3,512\end{array}\right] \times 2$} & \multicolumn{1}{c}{$\left[\begin{array}{l}3 \times 3,512 \\ 3 \times 3,512\end{array}\right] \times 2$}\\ \cline{1-7}
\multicolumn{1}{c|}{conv6}                  & \multicolumn{6}{c}{$\left[\begin{array}{c}   3 \times 3, stride~1, padding~1 \\ batch~norm \\ ReLU \end{array}\right]$}   \\ \cline{1-7}
\multicolumn{1}{c|}{up5}                    & \multicolumn{5}{c|}{$\left[\begin{array}{c} conv~3 \times 3, stride~1, padding~1 \\ batch~norm \\ ReLU \\ bilinear~interpolation \end{array}\right]$} \\ \cline{1-7}
\multicolumn{1}{c|}{up4}                    & \multicolumn{5}{c|}{$\left[\begin{array}{c} conv~3 \times 3, stride~1, padding~1 \\ batch~norm \\ ReLU \\ bilinear~interpolation \end{array}\right]$} \\ \cline{1-7}
\multicolumn{1}{c|}{up3}                    & \multicolumn{5}{c|}{$\left[\begin{array}{c}  conv~3 \times 3, stride~1, padding~1 \\ batch~norm \\ ReLU \\ bilinear~interpolation\end{array}\right]$} & \multicolumn{1}{c}{$\left[\begin{array}{c}  conv~3 \times 3, stride~1, padding~1 \\ batch~norm \\ Leaky~ReLU \\ bilinear~interpolation\end{array}\right]$} \\ \cline{1-7}
\multicolumn{1}{c|}{up2}                    & \multicolumn{5}{c|}{$\left[\begin{array}{c}  conv~3 \times 3, stride~1, padding~1 \\ batch~norm \\ ReLU \\ bilinear~interpolation \end{array}\right]$} & \multicolumn{1}{c}{$\left[\begin{array}{c}  conv~3 \times 3, stride~1, padding~1 \\ batch~norm \\ Leaky~ReLU \\ bilinear~interpolation\end{array}\right]$} \\ \cline{1-7}
\multicolumn{1}{c|}{up1}                    & \multicolumn{5}{c|}{$\left[\begin{array}{c} \\ conv~3 \times 3, stride~1, padding~1 \\ batch~norm \\ ReLU \\ bilinear~interpolation\end{array}\right]$} & \multicolumn{1}{c}{$\left[\begin{array}{c} conv~3 \times 3, stride~1, padding~1 \\ batch~norm \\ Leaky~ReLU \\ bilinear~interpolation \end{array}\right]$}\\ \cline{1-7}
\multicolumn{1}{c|}{conv7}                  & \multicolumn{5}{c|}{$\left[\begin{array}{c}   3 \times 3, stride~1, padding~1 \\ batch~norm \\ ReLU \end{array}\right]$} & \multicolumn{1}{c}{$\left[\begin{array}{c}   3 \times 3, stride~1, padding~1 \\ batch~norm \\ Leaky~ReLU\end{array}\right]$} \\ \cline{1-7}
\multicolumn{1}{c|}{conv8}                  & \multicolumn{5}{c|}{ $1\times 1, stride~1, padding~0$} & \multicolumn{1}{c}{$1\times 1, stride~1, padding~0$} \\ \cline{1-7}
             
\end{tabular}
\end{adjustbox}
\end{center}

\end{table*}

\clearpage
%
%
\bibliographystyle{splncs04}
\bibliography{egbib}